\documentclass[pmlr]{jmlr}

\RequirePackage{graphicx}
 \usepackage{booktabs}
\usepackage{longtable}
\usepackage{microtype}
\usepackage{graphicx}
\usepackage{subcaption}
\usepackage{multirow}
\usepackage{adjustbox}
\usepackage{booktabs} 
\usepackage{tikz}
\usepackage{amsmath}
\usepackage{hyperref}
\usepackage{enumitem}
\usepackage{csquotes}
\usepackage{amssymb}
\usepackage{mathtools}
\usepackage{bm} 

\makeatletter
\def\set@curr@file#1{\def\@curr@file{#1}} 
\makeatother
\usepackage[load-configurations=version-1]{siunitx} 

\theorembodyfont{\upshape}
\theoremheaderfont{\scshape}
\theorempostheader{:}
\theoremsep{\newline}

\title[GRAFT: Decoupling  Ranking and Calibration for Survival Analysis]{GRAFT: Decoupling  Ranking and Calibration for Survival Analysis}

\author{\Name{Mohammad Ashhad}
       \Email{mohammad.ashhad@kaust.edu.sa}\\ 
       \addr BESE\\
       KAUST\\
       Thuwal, KSA 
       \AND
       \Name{Robert Hoehndorf}
       \Email{robert.hoehndorf@kaust.edu.sa}\\ 
       \addr CEMSE\\
       KAUST\\
       Thuwal, KSA
       \AND
       \Name{Ricardo Henao}
       \Email{ricardo.henao@duke.edu}\\ 
       \addr Dept. of Bioinformatics \& Biostatistics\\
       Duke University\\
       Durham, USA} 

\begin{document}

\maketitle

\begin{abstract}
Survival analysis is complicated by censored data, high-dimensional features, and non-linear interactions. Classical models offer interpretability and superior calibration but are restricted to linear or predefined functional forms, while deep learning models are flexible and achieve strong discriminative performance, but tend to produce poorly calibrated survival estimates. To address this trade-off, we propose GRAFT (Gated Residual Accelerated Failure Time), a novel AFT model that decouples prognostic ranking from survival calibration. GRAFT's hybrid architecture combines a linear AFT model with a non-linear residual neural network, and it also integrates stochastic gates for automatic feature selection. The model is trained by optimizing a differentiable, C-index-aligned ranking loss using stochastic conditional imputation from local Kaplan-Meier estimators, while calibrated survival estimates are obtained through simple post-training calibration. In public benchmarks, GRAFT outperforms baselines in discrimination and calibration, while remaining robust and sparse in high-noise settings. 
\end{abstract}

\section{Introduction} 
\label{sec:introduction} 
Survival analysis, or time-to-event modeling, is a foundational branch of statistics focused on analyzing the expected duration of time until one or more events occur \citep{clark2003}. This methodology is indispensable in fields where the ``when'' is as critical as the ``if''. In clinical medicine, it is used to predict the prognosis of the patient \citep{he2023review}, assess the efficacy of new treatments by modeling the time to relapse \citep{kassani2015survival}, and identify risk factors for mortality \citep{abbasi2024survival}. In engineering, it serves as a reliability analysis by predicting the time-to-failure of mechanical components \citep{meeker2014statistical}. In finance, it is used to model the time-to-default in credit risk scoring \citep{dirick2017time}.

The principal challenge that distinguishes survival analysis from standard regression is the presence of censoring. A subject is right-censored if the event of interest has not occurred by the end of the study period, or if the subject is lost to follow-up. Standard regression techniques, such as Ordinary Least Squares (OLS), would yield biased estimates if applied to these types of partially observed data. Consequently, a specialized set of models has been developed to handle censoring appropriately \citep{clark2003}.

For decades, the field has been dominated by two classical approaches. The first is the non-parametric Kaplan-Meier (KM) estimator \citep{kaplan1958}, which provides a simple estimate of the (marginal) survival function but cannot incorporate features or covariates. The second and most prominent is the Cox Proportional Hazards (CoxPH) model \citep{cox1972}. As a semi-parametric model, CoxPH is highly valued for its calibration and its interpretable log-linear relationship between features and a {\em hazard ratio}. However, its validity rests on the strict Proportional Hazards (PH) assumption that the effect of a covariate on the hazard is constant over time. This assumption is frequently violated in real-world data. Parametric Accelerated Failure Time (AFT) models \citep{saikia2017review} offer an alternative by assuming a direct, linear relationship between covariates and the logarithm of survival time, but, in turn, they are limited by rigid distributional assumptions ({\em e.g.}, Weibull or Log-Normal). 

The advent of large-scale, complex and noisy datasets such as high-dimensional genomic data or granular electronic health records (EHRs) has further exposed the limitations of these classical models. The high-dimensionality of modern datasets introduces a critical need for robust feature selection. Although methods like LASSO (L1 regularization) can be applied to CoxPH models as a feature selection mechanism \citep{tibshirani1997}, they exhibit selection instability in the presence of correlated predictors \citep{zou2005, zhao2006}. Beyond feature selection, the non-linear and complex interactions present in modern datasets have spurred the development of machine learning and deep learning approaches, such as Random Survival Forests \citep[RSF,][]{ishwaran2008} and various deep neural networks such as DeepSurv \citep{katzman2018}. Although these models often achieve superior discriminative performance by capturing complex patterns, they tend to produce poorly calibrated survival estimates. Deep learning models, in particular, are also prone to overfitting and can be sensitive to irrelevant or ``noisy'' features.

To bridge the gap between the calibration strength of classical models and the discriminative power of deep learning, we propose the \textbf{Gated Residual AFT (GRAFT)} model. A key design principle of GRAFT is the explicit decoupling of prognostic ranking from survival calibration. Our model is a novel AFT architecture designed to be flexible, robust, and sparse simultaneously. It introduces a hybrid framework that learns both linear and non-linear feature effects, while employing a stochastic gating mechanism to perform automatic feature selection during training.
The primary contributions of this work are threefold.
\begin{enumerate}[itemsep=0pt,topsep=0pt,leftmargin=4mm]
\item \textbf{Decoupling of Ranking and Calibration:} We propose an explicit separation between prognostic ranking and survival calibration. GRAFT is trained to optimize discrimination via a differentiable ranking objective, while calibrated survival estimates are obtained through a simple post-training calibration step.
\item \textbf{A Hybrid AFT Architecture:} We combine a linear AFT model with a non-linear residual MLP, allowing the model to capture both simple linear effects and complex non-linear interactions. The architecture is trained using stochastic conditional imputation from local Kaplan-Meier estimators to handle censored observations, and optimized by directly minimizing a differentiable soft-ranking loss via Monte Carlo averaging, in alignment with the concordance index \citep[C-index,][]{harrell1982}.
\item \textbf{Integrated Stochastic Feature Selection:} We incorporate Stochastic Gates (STG) \citep{yamada2020}, a differentiable feature selection mechanism based on continuous relaxation of Bernoulli variables, directly into the model architecture. This enables the model to automatically identify relevant features and suppress noisy or redundant inputs during training, providing robustness in high-dimensional settings.
\end{enumerate}

We validate our model on six public benchmark datasets (GBSG, METABRIC, SUPPORT, NWTCO, FLCHAIN, and AIDS), demonstrating that GRAFT consistently outperforms both classical and modern baselines in both ranking and calibration. We further conduct experiments with injected noise features, proving that our model's stochastic gating mechanism successfully identifies and discards irrelevant information, a critical capability that traditional models lack.

\subsection*{Generalizable Insights about Machine Learning in the Context of Healthcare}
\begin{itemize}
\item \textbf{Decoupling ranking and calibration leads to improvements in survival analysis.} Deep learning survival models typically couple discrimination and calibration into a single objective, forcing a trade-off between the two. Our experiments demonstrate that explicitly separating these objectives, optimizing ranking during training, and post-hoc calibrating survival estimates yield models that are simultaneously better at discrimination and calibration, both of which are critical for clinical decision-making.
\item \textbf{Integrated feature selection is critical for robustness in healthcare scenarios.} Real-world clinical datasets are frequently high-dimensional, containing irrelevant measurements, noisy biomarkers, and spurious correlations. Our experiments demonstrate that models without explicit feature selection mechanisms degrade significantly under such conditions, while GRAFT's stochastic gating consistently identifies and suppresses uninformative features.
\end{itemize}

\section{Related Work}
\label{sec:related_work}
The methodologies for survival analysis can be broadly categorized into classical statistical models and modern machine learning approaches. Our proposed model, GRAFT, draws inspiration from both traditions.

{\bf Classical Survival Models} For decades, the field of survival analysis has been dominated by a few robust and interpretable models. The most foundational of these is the non-parametric Kaplan-Meier (KM) estimator, which provides a population-level estimate of the survival function $S(t)$ from time-to-event data \citep{kaplan1958}. Although essential for summarizing survival data, the KM estimator cannot incorporate individual-specific covariates to provide personalized predictions. To address this limitation, the Cox Proportional Hazards (CoxPH) model was introduced \citep{cox1972}. The CoxPH model is a semi-parametric method that estimates the effect of covariates $x$ on the hazard rate, $\lambda(t|x)$, at time $t$ without making assumptions about the underlying shape (or distribution) of the baseline hazard, $\lambda_0(t)$:
\begin{align}\label{eq:coxph}
\lambda(t|x) = \lambda_0(t) \exp(x^T \beta) \,,
\end{align}
The output of the CoxPH model, the hazard ratio $\exp(\beta)$, is highly interpretable and has made it the {\em de facto} standard in clinical research. However, it relies on the strict Proportional Hazards (PH) assumption that the effect of covariates is constant over time, which is often violated in practice \citep{clark2003}.

In contrast, parametric models such as the Weibull model, which can be formulated as an Accelerated Failure Time (AFT) model, constitute an alternative \citep{kalbfleisch2002statistical}. AFT models assume that covariates act to rescale time directly, such as by accelerating or decelerating the time-to-event. Although fully parametric models can be more efficient if the underlying distribution is correctly specified, their rigidity is a significant limitation. The proposed GRAFT model is built upon the flexible AFT framework, but, as we show, it circumvents the need for strict distributional assumptions.

{\bf Deep Learning for Survival Analysis} With the rise of high-dimensional data, traditional models that assume linearity (like CoxPH and AFT) have proven insufficient for capturing complex, non-linear patterns. This has led to the development of deep learning models for survival analysis. Although these models often achieve strong discriminative performance, a consistent finding in benchmark studies is that their calibration tends to be poor compared to classical approaches \citep{burk2024large}.

One of the most prominent early-proposed models is DeepSurv \citep{katzman2018}. DeepSurv directly adapts the CoxPH model to a deep learning context by replacing the linear term $x^T \beta$ in \eqref{eq:coxph} with a deep feed-forward neural network $h_{\theta}(x)$, and optimizes a modified Cox partial log-likelihood loss. While effective at modeling non-linear covariate effects, DeepSurv still inherits the limitations of the proportional hazards assumption.

Other approaches have sought to discretize time and reformulate the problem, enabling more flexible hazard modeling. DeepHit \citep{lee2018} is a landmark model in this area. The model discretizes the time horizon into a set of bins and learns the joint distribution of survival time and event type, making it a natural fit for competing risks. By optimizing a loss function that combines a log-likelihood for the observed event time and a rank-based loss for censored subjects, DeepHit can estimate the full survival probability curve without being restricted by the PH assumption. DeepHit combines a log-likelihood term for calibration and a ranking loss for discrimination into a single objective, controlled by a scalar trade-off parameter, coupling the two objectives.

GRAFT addresses the key limitations of existing deep learning approaches through two design choices. First, it explicitly decouples discrimination from calibration: the model is trained to optimize ranking, while calibrated survival estimates are obtained through a dedicated post-training step. Second, it integrates stochastic gates directly into the architecture for automatic feature selection, providing robustness to noise and overfitting in high-dimensional settings. Together, these choices position GRAFT as a model that combines the discriminative power of deep learning with the calibration reliability of classical approaches.

\section{The Gated Residual AFT Model}
\label{sec:methodology} 
The proposed model, Gated Residual AFT (GRAFT), is a survival model that integrates local nonparametric imputation, stochastic feature selection, and a non-linear residual MLP.
The model is trained via minibatch stochastic gradient descent by directly optimizing a differentiable approximation of the rank-correlation between event times and model predictions using Monte Carlo averaging as explained in the following.

\subsection{Problem Formulation}
\label{sec:problem_formulation}
We consider a dataset $D = \{(x_i, t_i, \delta_i)\}_{i=1}^N$, where $N$ is the number of subjects. For each subject $i$, $x_i \in \mathbb{R}^p$ is a vector of $p$ baseline features, $t_i > 0$ is the observed event time and $\delta_i \in \{0, 1\}$ is the event indicator, with $\delta_i = 1$ denoting an observed event and $\delta_i = 0$ denoting right-censoring, {\em i.e.}, $t_i=\min(y_i,c_i)$ and $\delta_i=\mathbb{I}(t_i<c_i)$ given the true event time and the censoring time $y_i$ and $c_i$, respectively, of which only one is observed as $t_i$; and $\mathbb{I}(\cdot)$ is the indicator function.

The classical Accelerated Failure Time (AFT) model \citep{kalbfleisch2002statistical} provides a linear foundation for survival analysis by letting
\begin{align}
\log(T_i) = x_i^T \beta + \mu + \epsilon_i , \label{eq:aft}
\end{align}
where $\beta \in \mathbb{R}^p$ is the vector of coefficients, $\mu$ is an intercept term, and $\epsilon_i$ is an error term. The challenge is that $\log(T_i)$ is unknown for subjects with censoring ($\delta_i=0$), which AFT models circumvent by specifying the distribution of $\epsilon_i$ in the location-scale family so the loss for censored observations can be written in closed form. Moreover, standard AFT models assume a linear relationship between features and log-survival event time, which may be insufficient for capturing complex, non-linear interactions in high-dimensional survival data.

\subsection{GRAFT Model Overview}
\label{sec:model_overview}
Instead of directly modeling $\log(T_i)$ as in \eqref{eq:aft}, GRAFT learns a prognostic score $s_i \in \mathbb{R}$ that is monotonic with the log-survival time.
Specifically,
\begin{align}
\tilde{x}_i &= g_{\eta,i} \odot x_i \label{eq:gated_input}\\
\phi_i &= \tilde{x}_i + f_{\theta}(\tilde{x}_i) \label{eq:gated_features}\\
s_i &= \beta^T \phi_i + \mu , \label{eq:score}
\end{align}
where $\odot$ denotes element-wise product and $g \in (0, 1)^p$ in \eqref{eq:gated_input} is a $p$-dimensional vector with values sampled from a stochastic gate parameterized by $\eta\in\mathbb{R}^p$ used here to mask the input features $x_i$.
During training, the gate values are sampled from a binary distribution using the reparameterization trick.
Note that since the gate vector parameter $\eta$ is shared between all subjects, it enables feature selection at the population level.
This gating mechanism is detailed in Section~\ref{sec:stg}.

The gated features in \eqref{eq:gated_input} are then fed through a non-linear (residual) MLP $f_{\theta}: \mathbb{R}^p \rightarrow \mathbb{R}^p$ with parameters $\theta$.
For simplicity, in our experiments we use a fully-connected network with two layers, tanh activation function and $d_h$ hidden units; however, deeper architectures or alternative parameterizations can also be considered.

The score $s_i$ is obtained in \eqref{eq:score} as a linear combination with parameters $\beta \in \mathbb{R}^p$ and $\mu \in \mathbb{R}$ of the gated features $\tilde{x}_i$ and the residual MLP $f_{\theta}(\tilde{x}_i)$, thus the representation $\phi_i$ in \eqref{eq:gated_features} captures both the direct effect of selected features and their non-linear interactions learned by the residual network.

The complete set of model parameters is $\Psi = \{\eta, \theta, \beta, \mu\}$, denotes the parameters of the stochastic gate described in Section~\ref{sec:stg}.
All parameters are learned jointly during training.
In the following, we detail the stochastic imputation procedure for handling censored observations in Section~\ref{sec:local_km}, the stochastic gate mechanism in Section~\ref{sec:stg}, the training objective in Section~\ref{sec:loss}, and post-training survival curve estimation in Section~\ref{sec:survival_estimation}).

\subsection{Stochastic Conditional Imputation via Local KM}
\label{sec:local_km}
To train our AFT model on censored data, we require target values representing the underlying survival times.
Instead of assuming a parametric distribution for $\epsilon_i$ in \eqref{eq:aft} and using its conditional density function to model censored data as in standard AFT modeling (see \citet{kalbfleisch2002statistical} for details), or using an imputation mechanism based on point estimates, we employ a {\em stochastic conditional imputation} approach based on local and nonparametric survival estimation \citep{dabrowska1987non}.
This is done to preserve the distributional uncertainty inherent in censored data by letting the model train with plausible event-time realizations rather than a single imputed value.
However, we still make the standard assumption in survival analysis that given the input features $X$, the true event time $Y$ and the censoring time $C$ are independent, {\em i.e.}, $Y \perp\!\!\!\perp C \mid X$.

{\bf Local KM Estimation.}
For each censored observation $i$, \textit{i.e.}, with $\delta_i = 0$, we 
construct a local neighborhood $\mathcal{V}(x_i) \subseteq \{1, \ldots, N\}$ 
in the input feature space (using the Euclidean distance with standardized 
features) by expanding outward from $x_i$ until $k$ neighbors with observed 
events ($\delta_j = 1$) are found; all neighbors encountered during this 
expansion, including censored ones, are retained in $\mathcal{V}(x_i)$, so 
$|\mathcal{V}(x_i)| \geq k$. We set $k = 10$.


Using only the subjects in $\mathcal{V}(x_i)$, we fit a Kaplan-Meier (KM) survival estimator \citep{kaplan1958}, denoted as $\hat{S}(t \mid x_i)$ to approximate the conditional survival function $S(t \mid x_i)$. 
Since we know that the subject $i$ has survived beyond their censoring time $t_i$, we condition the survival distribution on this information:
\begin{equation}
\hat{S}(t \mid x_i, T_i > t_i)
= \frac{\hat{S}(t \mid x_i)}{\hat{S}(t_i \mid x_i)},
\quad t > t_i .
\end{equation}
This conditional survival function is a nonparametric estimate of the distribution of $T_i$ given covariates $x_i$ and survival beyond $t_i$.
We use such a $k$-nearest-neighbor-based estimator for $S(t \mid x_i)$ because it is an instance of the Beran conditional product-limit estimator, which is consistent under conditional independent censoring \citep{beran1981,dabrowska1989,dabrowska1987non}.

From a practical perspective, it is worth noting that the neighborhoods and corresponding local KM curves are computed once using the full training set prior to optimization, defining fixed conditional imputation distributions that are used throughout training.

{\bf Stochastic Sampling.}
Rather than summarizing the conditional survival distribution $S(t \mid x_i)$ to a single point estimate, {\em e.g.}, its expectation, we follow the principle of multiple imputation \citep{rubin1987}, and draw samples directly from the conditional distribution.
For uncensored subjects ($\delta_i = 1$), we simply use the observed log-time $Y_i^* = \log(t_i)$.
For censored subjects ($\delta_i = 0$), we sample imputed event times using
\begin{equation}
T_i^* \sim \hat{F}(\cdot \mid x_i, T_i > t_i), \label{eq:km_sampling}
\end{equation}
where $\hat{F}(t \mid x_i, T_i > t_i) = 1 - \hat{S}(t \mid x_i, T_i > t_i)$ is the conditional cumulative distribution function and then set $Y_i^* = \log(T_i^*)$.
In practice, this is done by inverse-transform sampling \citep{devroye2006nonuniform}.
During training, multiple independent imputation values ($M=5$) are generated for censored subjects within each minibatch, allowing the learning objective to account for the uncertainty of imputation through Monte Carlo averaging, as detailed in Section~\ref{sec:loss}.

\subsection{Feature Selection with Stochastic Gates}
\label{sec:stg}
To achieve sparsity and identify relevant features in noisy or high-dimensional settings, we integrate Stochastic Gates (STG) \citep{yamada2020}, directly into the model architecture. STG uses a continuous relaxation of Bernoulli random variables based on Gaussian distributions to learn a probabilistic binary mask over the input features.

For the $j$-th feature index (where $j \in \{1, \dots, p\}$), we follow the generating process
\begin{align}
    \epsilon_i & \sim \mathcal{N}(0, \sigma^2) \notag \\
    g_{\eta_j,i} & = \max\left(0, \min\left(1, (\eta_j + \epsilon_i\right)\right)) , \label{eq:stg_clamp}
\end{align}
where $\mathcal{N}(0, \sigma^2)$ is a Gaussian distribution with variance $\sigma^2$, $\eta_j$ is an element of $\eta \in \mathbb{R}^p$, which denotes the learnable mean gate parameter, and $g_{\eta_j,i}\in(0,1)$ is a gate value sample for component $j$ of $x_i$.
These are shared by all data points in a minibatch to ensure that all subjects are evaluated with the same feature mask, thereby learning population-level feature importance rather than subject-specific patterns.
The mechanism in \eqref{eq:stg_clamp} uses the reparameterization trick \citep{miller2017reducing}, which allows gradient-based optimization of $\eta$.

During inference ({\em i.e.}, at test time) and for convenience, the stochastic sampling in \eqref{eq:stg_clamp} is replaced by the (deterministic) expected value of the gate, {\em i.e.},
\begin{equation}
g_{\eta_j} = \max\left(0, \min\left(1, (\eta_j\right)\right)) ,\label{eq:stg_deterministic}
\end{equation}
We use deterministic gates during inference to eliminate sampling variance, provide reproducible predictions, and reduce computational overhead due to prediction averaging \citep{yamada2020}; which in some cases may lead to more robust results.

To encourage sparsity, we add a differentiable approximation of the $\ell_0$ norm to the loss function, which penalizes the expected number of active (non-zero) gates:
\begin{equation}
\mathcal{L}_{\text{reg}} = \lambda_{\text{L0}} \sum_{j=1}^p \mathbb{E}[g_j > 0] = \lambda_{\text{L0}} \sum_{j=1}^p \Phi\left(\frac{\eta_j}{\sigma}\right) , \label{eq:stg_reg}
\end{equation}
where $\lambda_{\text{L0}} > 0$ is a hyperparameter controlling the strength of the sparsity penalty (we set $\lambda_{\text{L0}}=0.01$), and $\Phi(\cdot)$ denotes the cumulative distribution function of the standard Gaussian distribution.
The expectation $\mathbb{E}[g_j > 0] = P(\eta_j + \epsilon_j > 0) = \Phi(\eta_j / \sigma)$ follows from the properties of the Gaussian distribution \citep{yamada2020}. 

\subsection{Training Objective: Differentiable Ranking Loss}
\label{sec:loss}
We train GRAFT by optimizing a ranking-based objective that aligns with the concordance index (C-index) \citep{harrell1982}, commonly used as a evaluation metric in survival analysis (with discussed below).
Our loss function combines a differentiable approximation of Spearman's rank correlation with the STG penalty in \eqref{eq:stg_reg} and standard weight regularization.

{\bf Differentiable Ranking Loss.}
We optimize the negative Spearman's rank correlation between the predicted scores and target log-times for each minibatch of size $m$, with censored events imputed as described in Section~\ref{sec:local_km}, to encourage predictions to match the order of the target event times.
Since the ranking operator needed by the Spearman's rank correlation is non-differentiable, we use the differentiable soft-ranking operator of \citet{blondel2020}, which provides a continuous relaxation to the rank function while preserving the ordering of the input scores.
This operator is computed through isotonic regression using the Pool Adjacent Violators (PAVA) algorithm \citep{best2000minimizing}.

For a minibatch $\mathcal{B} = \{i_1, \ldots, i_m\}$ with predicted scores $s_{\mathcal{B}} \in \mathbb{R}^m$ and imputed targets $Y^*_{\mathcal{B}} \in \mathbb{R}^m$, the ranking loss is
\begin{equation}
\mathcal{L}_{\text{rank}}(s_{\mathcal{B}}, Y^*_{\mathcal{B}}) = - \frac{\text{cov}(\hat{R}_{\tau}(s_{\mathcal{B}}), R(Y^*_{\mathcal{B}}))}{\sigma_{\hat{R}_{\tau}(s_{\mathcal{B}})} \cdot \sigma_{R(Y^*_{\mathcal{B}})}} , \label{eq:rank_loss}
\end{equation}
where $\hat{R}_{\tau}(s_{\mathcal{B}}) \in \mathbb{R}^m$ and $R(Y^*_{\mathcal{B}}) \in \mathbb{R}^m$ denote the soft-ranking vector of predicted scores and the (hard) ranking vector of imputed targets within the minibatch, respectively, $\text{cov}(\cdot, \cdot)$ and $\sigma(\cdot)$ denote covariance and standard deviation, respectively, and $\tau > 0$ in $\hat{R}_{\tau}(\cdot)$ controls the degree of smoothing of the relaxed soft rank function (we use $\tau = 0.1$). 

{\bf Monte Carlo Averaging over Imputations.}
To account for the uncertainty in imputed values for censored observations, we combine Monte Carlo averaging with minibatch stochastic gradient descent. For each minibatch $\mathcal{B}$, we generate $M=5$ independent imputation values by sampling from the pre-fitted conditional KM distributions (as described in Section~\ref{sec:local_km}). 

Specifically, for Monte Carlo draw $k \in \{1, \ldots, M\}$, we construct an imputed target vector $Y^{*(k)}_{\mathcal{B}} \in \mathbb{R}^m$. For uncensored subjects ($\delta_i = 1$), we set $Y_i^{*(k)} = \log(t_i)$ (fixed across all $k$). For censored subjects ($\delta_i = 0$), we sample $T_i^{*(k)} \sim \widehat{F}_i(\cdot \mid x_i, T_i > t_i)$ (re-sampled for each $k$ as in~\eqref{eq:km_sampling}) and set $Y_i^{*(k)} = \log(T_i^{*(k)})$.
The ranking loss in~\eqref{eq:rank_loss} then becomes the average over $M$ sets of imputed target events
\begin{equation}
\bar{\mathcal{L}}_{\text{rank}}(\mathcal{B}) = \frac{1}{M} \sum_{k=1}^M \mathcal{L}_{\text{rank}}(s_{\mathcal{B}}, Y^{*(k)}_{\mathcal{B}}) .  \label{eq:mc_rank_loss}
\end{equation}
This Monte Carlo averaging ensures that the model optimizes ranking consistency in expectation over the imputation distribution, rather than overfitting to any single imputed realization.

{\bf Complete Training Objective.}
The final loss function for each minibatch combines the Monte Carlo-averaged ranking loss from~\eqref{eq:mc_rank_loss}, the STG penalty from~\eqref{eq:stg_reg}, and standard $L_2$ weight regularization on all model parameters $\Psi = \{\eta, \theta, \beta, \mu\}$,
\begin{equation}
\mathcal{L}_{\text{total}}(\mathcal{B}) = \bar{\mathcal{L}}_{\text{rank}}(\mathcal{B}) + \mathcal{L}_{\text{reg}} + \alpha_{\text{L2}} \|\Psi\|_2^2 , \label{eq:total_loss}
\end{equation}
where $\alpha_{\text{L2}} > 0$ is the regularization hyperparameter (we use $\alpha_{\text{L2}}=10^{-4}$).
We minimize $\mathcal{L}_{\text{total}}$ using the Adam optimizer \citep{kingma2014adam} via minibatch stochastic gradient descent with a learning rate of $10^{-3}$ to minimize the expected soft-Spearman ranking loss, averaged over Monte Carlo draws from the fixed conditional imputation distributions.

Under the assumption of conditional independent censoring, {\em i.e.}, $Y \perp\!\!\!\perp C \mid X$, and standard regularity conditions for local Kaplan-Meier estimators (see \citet{dabrowska1989} for details), the estimated imputation distributions $\widehat{F}(\cdot \mid x_i, T_i > t_i)$ in \eqref{eq:km_sampling} are consistent estimators of the true conditional event-time law $T_i \mid x_i, T_i > t_i$.
Consequently, since the C-index \citep{harrell1982} quantifies the pairwise ranking performance, optimizing a differentiable rank-based surrogate constitutes a tractable objective to improve concordance under censoring, {\em i.e.}, improving rank correlation will also improve the C-index as long as $\widehat{F}(\cdot \mid x_i, T_i > t_i) \to F(T_i \mid x_i, T_i > t_i)$ (in probability). 

\subsection{Survival Function Estimation}
\label{sec:survival_estimation}
The GRAFT model, as described in Section~\ref{sec:model_overview}, produces a prognostic score $s_i \in \mathbb{R}$ that is monotonic with log-survival times and optimized for ranking, thus for concordance as discussed above.
However, it does not directly produce the survival function $\hat{S}(t \mid x) = \mathbb{P}(T > t \mid x)$.
To obtain survival function estimates from the learned GRAFT scores, we adopt a post-training calibration procedure using the popular CoxPH model \citep{cox1972}, which we specify as follows.
\begin{equation}\label{eq:post_cox}
\lambda(t \mid s) = \lambda_0(t)\exp(\beta_{\text{cox}} \cdot s),
\end{equation}
where $\lambda_0(t)$ denotes the baseline hazard function,  $\beta_{\text{cox}} \in \mathbb{R}$ is the regression coefficient associated with the GRAFT score obtained using~\eqref{eq:score}.
The CoxPH model is fit by maximizing the partial likelihood, yielding an estimate of $\beta_{\text{cox}}$ along with the standard Breslow estimator for the baseline cumulative hazard function $\Lambda_0(t) = \int_0^t \lambda_0(u)\,du$ \citep{lin2007breslow}.

For a test subject with input features $x_{te}$, we first compute the corresponding GRAFT score $s_{te}$ using~\eqref{eq:score}.
The estimated survival function is then given by
%

\begin{equation}
\hat{S}(t \mid x_{te})
= \exp\left(-\Lambda_0(t)\cdot\exp\left(\beta_{\text{cox}} \cdot s_{te}\right)\right)
\end{equation}
where $\Lambda_0(t)$ is evaluated at arbitrary time points through linear interpolation of the Breslow estimator obtained from the training data. 
This procedure leverages the GRAFT scores as a learned feature representation and uses CoxPH as a post-processing layer to convert these scores into time-dependent survival probability estimates.
Importantly, this step does not modify the GRAFT parameters $\Psi$ or its discrimination ability (C-index), and only requires fitting a one-dimensional CoxPH model, which incurs minimal additional computation, but guarantees calibrated predictions as long as $\Lambda_0(t)$ holds for $x_{te}$ \citep{lin2007breslow}.
This is not a strong assumption because one usually assumes that $\{(x_{te},t_{te})\}$ and ${\cal D}$ (the training dataset), are i.i.d., however, the assumption that the baseline hazard function $\lambda_0(t)$ is not dependent on covariates $x$ may not hold in practice, which can indeed affect calibration guarantees. We empirically investigate whether this assumption limits the calibration performance by comparing the CoxPH-based approach in \eqref{eq:post_cox} with non-parametric isotonic regression calibration \citep{tibshirani2011nearly}, in Appendix~\ref{sec:isotonic_calibration}.

\section{Experiments}
\label{sec:experiments}
All experiments were conducted on a system with an AMD Ryzen Threadripper PRO 5975WX CPU and NVIDIA RTX 5000 Ada GPU, with deep learning models trained on the GPU. Code for reproducing all experiments is available at: \url{https://github.com/anonymous-785-u/GRAFT}.

\subsection{Experimental Setup}
\label{sec:experimental_setup}

\textbf{Datasets:} We evaluate GRAFT on six public survival analysis benchmarks spanning diverse censoring rates and application domains. The GBSG (German Breast Cancer Study Group) dataset \citep{schumacher1994randomized} contains 2,232 breast cancer patients with 43\% censoring. The METABRIC dataset \citep{curtis2012genomic} provides genomic data for 1,904 breast cancer patients with 42\% censoring. The SUPPORT dataset \citep{knaus1995support} includes 8,873 critically ill hospitalized patients with 32\% censoring. The NWTCO dataset \citep{breslow1999design} comprises 4,028 pediatric kidney tumor patients with 86\% censoring. The FLCHAIN dataset \citep{dispenzieri2012use} contains 7,874 subjects from a serum free light chain study with 72\% censoring. Finally, the AIDS dataset \citep{hammer1997controlled} includes 1,151 HIV/AIDS patients with 92\% censoring. This collection spans low-censoring (32-43\%) and high-censoring (72-92\%) scenarios.

\textbf{Baselines:} We compare GRAFT with five representative survival models. Cox Proportional Hazards (CoxPH) \citep{cox1972} serves as the standard semi-parametric baseline. Weibull AFT \citep{wei1992accelerated} represents a fully parametric accelerated failure time model. DeepHit \citep{lee2018} is a deep learning model that discretizes time and learns joint distributions without assuming proportional hazards. DeepSurv \citep{katzman2018} adapts the Cox model to deep learning by using neural networks to learn non-linear covariate effects. The Random Survival Forest (RSF) \citep{ishwaran2008} provides an ensemble tree-based method.
The hyperparameters for all models are listed in Appendix~\ref{sec:hyperparams}. To validate that our ranking loss serves as a reliable training objective for concordance, we also record the training loss and test C-index at every epoch across all runs of GRAFT.

\textbf{Metrics and Evaluation:} We evaluate both ranking and calibration performance using two complementary metrics. Harrell's Concordance Index (C-index)~\citep{harrell1982} measures the probability that, for a random pair of subjects where one experiences the event before the other, the model correctly ranks their risk scores. Higher values indicate better discrimination, with the metric ranging from 0.5 (random) to 1.0 (perfect). The Integrated Brier Score (IBS) \citep{graf1999assessment} measures the mean squared error between predicted survival probabilities and observed outcomes over time, weighted by the inverse probability of censoring (IPCW). Lower values indicate better calibration, with the metric ranging from 0 (perfect) to 1 (worst).

Our evaluation employs 3-fold cross-validation with three random seeds to provide comprehensive uncertainty quantification. We report results using dual averaging perspectives. \textit{Fold-averaged results} are computed by averaging metrics across folds within each seed, then reporting the standard deviation across seeds, which captures variance from random initialization and stochastic training. \textit{Seed-averaged results} are computed by averaging metrics across seeds within each fold and then reporting the standard deviation over folds, which captures variance from data partitioning. Together, these two perspectives reveal both training stochasticity and sensitivity to train-test splits.


\subsection{Ablation Study}
\label{sec:ablation_intro}

To isolate the contributions of GRAFT's key architectural components, we evaluate three model variants. {\em Full GRAFT} represents the complete model with both stochastic gates and residual MLP. The {\em No STG} variant removes stochastic feature selection while retaining the residual MLP, using all input features without gating. The {\em Linear Only} variant removes both gates and MLP, reducing to a pure linear AFT model. We compare these variants by augmenting each dataset with additional noise covariates. For a dataset with $p$ original features, we add $kp$ noise features with multipliers $k \in \{3, 5, 7, 10\}$, where each noise feature is independently sampled from a standard Gaussian distribution. This experimental design allows us to quantify the contribution of the stochastic gates (by comparing {\em Full GRAFT} to {\em No STG}) and the residual MLP (by comparing {\em No STG} to {\em Linear Only}) to the overall discrimination and calibration performance.

\subsection{Noise Robustness Analysis}
\label{sec:noise_robustness}

We conduct a comprehensive noise robustness analysis by stress-testing all six models with additional noise covariates. For a dataset with $p$ original features, we augment it with $kp$ additional noise features at multipliers $k \in \{3, 5, 7, 10\}$, where each noise feature is independently sampled from a Student's t-distribution with 2 degrees of freedom. This distribution has infinite variance and generates extreme outliers, representing a more challenging and realistic scenario than Gaussian noise to assess feature selection robustness. This simulates the presence of spurious biomarkers or measurement artifacts in the datasets. By comparing all six models under identical noise conditions—where only the original $p$ features are truly predictive, we can directly evaluate the effectiveness of GRAFT's stochastic gating mechanism relative to the implicit regularization strategies employed by classical models (L2 penalties in CoxPH and Weibull, bootstrap aggregation in RSF) and the vulnerabilities of deep learning approaches without explicit feature selection (DeepHit, DeepSurv).

\subsection{Imputation Neighborhood Sensitivity Analysis} 

To assess the robustness of GRAFT's local KM imputation to neighborhood construction choices, we conduct a sensitivity analysis over three design dimensions: $i$) the number
of observed events required within a neighborhood $k$ $\in \{5, 10, 20, 30\}$; $ii$) the distance metric used to identify nearest neighbors, Euclidean (Euc) or Mahalanobis (Mah); and $iii$) the neighborhood composition, where Events-Only (EO) restricts the KM estimator to neighbors with observed events only, while Events+Censored (EC) includes all neighbors regardless of
censoring status. All other hyperparameters, the model architecture and the training procedure, are maintained as in Section \ref{sec:baseline_comparison}. Each configuration is evaluated using the same 3-fold cross-validation with 3 random seeds as the main experiment, and results are reported as mean C-index and IBS across runs.

\begin{table*}[t]
\centering
\caption{\small{Comparison of GRAFT and baselines across six datasets. Values reported as Mean (Fold-Std, Seed-Std) based on 3-fold cross-validation with 3 random seeds. Fold-averaged standard deviation represents variance from model initialization; seed-averaged standard deviation represents variance from data splitting. Best values per dataset in \textbf{bold}.}}
\label{tab:main_results}
\begin{adjustbox}{max width=\textwidth}
\small
\begin{tabular}{@{}llcccccc@{}}
\toprule
\textbf{Metric} & \textbf{Dataset} & \textbf{GRAFT} & \textbf{CoxPH} & \textbf{Weibull} & \textbf{DeepHit} & \textbf{DeepSurv} & \textbf{RSF} \\
\midrule
\multicolumn{8}{l}{\textit{Low Censoring Datasets}} \\
\midrule
C-Index & GBSG & \textbf{0.6730} (0.0008, 0.0022) & 0.6628 (0.0015, 0.0049) & 0.6630 (0.0015, 0.0051) & 0.6592 (0.0071, 0.0047) & 0.6629 (0.0024, 0.0050) & 0.6719 (0.0025, 0.0025) \\
(higher) & METABRIC & \textbf{0.6460} (0.0016, 0.0046) & 0.6332 (0.0020, 0.0097) & 0.6333 (0.0020, 0.0099) & 0.6211 (0.0050, 0.0199) & 0.6280 (0.0040, 0.0052) & 0.6431 (0.0016, 0.0100) \\
& SUPPORT & \textbf{0.6143} (0.0010, 0.0005) & 0.5686 (0.0011, 0.0041) & 0.5675 (0.0011, 0.0041) & 0.5737 (0.0058, 0.0023) & 0.6071 (0.0016, 0.0053) & 0.6057 (0.0005, 0.0048) \\
\midrule
IBS & GBSG & \textbf{0.1760} (0.0002, 0.0012) & 0.1814 (0.0003, 0.0015) & 0.1821 (0.0003, 0.0017) & 0.1828 (0.0006, 0.0004) & 0.1818 (0.0007, 0.0023) & 0.1767 (0.0006, 0.0006) \\
(lower) & METABRIC & 0.1630 (0.0007, 0.0045) & \textbf{0.1628} (0.0006, 0.0041) & 0.1636 (0.0007, 0.0037) & 0.1730 (0.0007, 0.0059) & 0.1728 (0.0024, 0.0042) & 0.1654 (0.0004, 0.0052) \\
& SUPPORT & \textbf{0.1891} (0.0005, 0.0018) & 0.2059 (0.0003, 0.0019) & 0.2071 (0.0003, 0.0019) & 0.2021 (0.0005, 0.0020) & 0.1956 (0.0005, 0.0021) & 0.1902 (0.0001, 0.0015) \\
\midrule
\multicolumn{8}{l}{\textit{High Censoring Datasets}} \\
\midrule
C-Index & NWTCO & \textbf{0.7173} (0.0012, 0.0055) & 0.7104 (0.0005, 0.0049) & 0.7113 (0.0006, 0.0047) & 0.6952 (0.0086, 0.0045) & 0.7101 (0.0011, 0.0075) & 0.6921 (0.0025, 0.0092) \\
(higher) & FLCHAIN & \textbf{0.7965} (0.0004, 0.0010) & 0.7689 (0.0378, 0.0400) & 0.7954 (0.0004, 0.0025) & 0.7695 (0.0018, 0.0181) & 0.7895 (0.0003, 0.0019) & 0.7917 (0.0004, 0.0047) \\
& AIDS & 0.7301 (0.0039, 0.0244) & \textbf{0.7635} (0.0080, 0.0181) & 0.7633 (0.0077, 0.0191) & 0.7012 (0.0263, 0.0113) & 0.6996 (0.0193, 0.0104) & 0.7593 (0.0025, 0.0082) \\
\midrule
IBS & NWTCO & \textbf{0.1016} (0.0003, 0.0007) & 0.1044 (0.0000, 0.0015) & 0.1064 (0.0000, 0.0017) & 0.1042 (0.0011, 0.0016) & 0.1073 (0.0002, 0.0016) & 0.1043 (0.0003, 0.0026) \\
(lower) & FLCHAIN & \textbf{0.0919} (0.0001, 0.0004) & 0.0961 (0.0056, 0.0060) & 0.0927 (0.0003, 0.0007) & 0.0972 (0.0005, 0.0016) & 0.0966 (0.0013, 0.0008) & 0.0928 (0.0003, 0.0009) \\
& AIDS & 0.0572 (0.0004, 0.0032) & \textbf{0.0556} (0.0003, 0.0031) & 0.0557 (0.0003, 0.0032) & 0.0692 (0.0062, 0.0039) & 0.0649 (0.0037, 0.0012) & 0.0559 (0.0006, 0.0031) \\
\bottomrule
\end{tabular}
\end{adjustbox}
\end{table*}

\section{Results and Discussion}
\label{sec:results}

\subsection{Baseline Comparison and Loss to C-index Alignment}
\label{sec:baseline_comparison}

Table~\ref{tab:main_results} shows that GRAFT outperforms baselines in five of six datasets in C-index and IBS. In low-censoring datasets, GRAFT demonstrates clear advantages. For example, on SUPPORT, GRAFT achieves C-index 0.6143 and IBS 0.1891, outperforming all baselines including DeepSurv (0.6071) and RSF (0.6057). A similar pattern can be observed in GBSG and METABRIC. In high-censoring datasets, GRAFT maintains strong performance despite sparse event information. In FLCHAIN (72\% censoring), GRAFT achieves a C-index of 0.7965 along with the best IBS of 0.0919. The consistent superiority over classical and deep learning models validates GRAFT's hybrid architecture and training objective, providing strong empirical evidence for its utility in survival analysis.

However, GRAFT shows a clear limitation on AIDS (92\% censoring), achieving C-index 0.7301 versus CoxPH's 0.7635. With only approximately 92 observed events, GRAFT's local neighborhoods contain insufficient information for stable KM curve estimation, and the added model complexity is not justified. This suggests a boundary condition: when event rates fall below approximately 10\%, practitioners should consider simpler parametric models such as CoxPH.
DeepHit and DeepSurv perform competitively in some datasets but lack integrated feature selection, making them vulnerable in high-dimensional, high noise settings (Section~\ref{sec:noise_robustness_results}). The dual standard deviation reporting reveals GRAFT's stable convergence (fold-averaged std: 0.0002-0.0039 on the C-index) despite jointly optimizing gates, MLP parameters, and linear coefficients. Seed-averaged standard deviations (0.0005-0.0244) reflect competitive data partitioning sensitivity, with AIDS showing the highest variance due to extreme censoring.

\begin{figure}[t]
    \centering
    \includegraphics[width=0.7\textwidth]{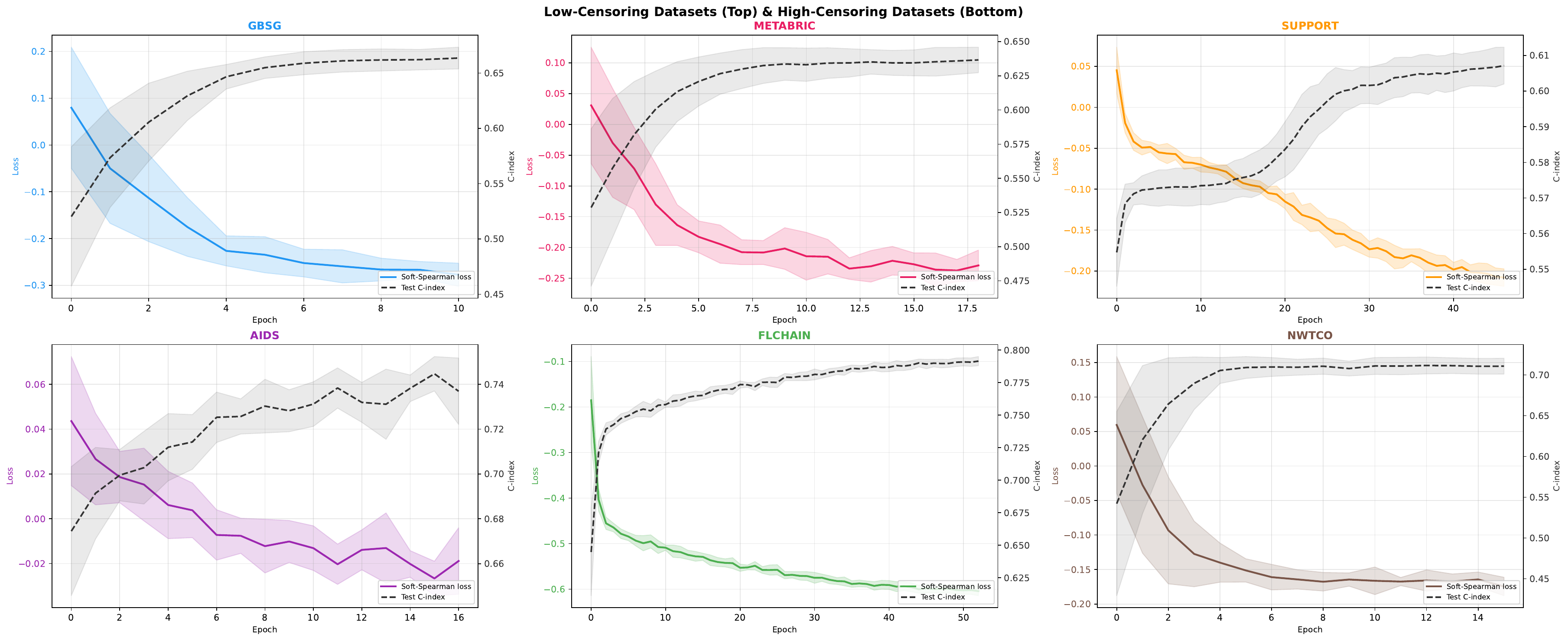}
    \caption{\small{GRAFT training dynamics across all six datasets. Each panel 
    shows the mean $\pm$ standard deviation of the soft-Spearman 
    loss and test C-index over training epochs across all runs.}}
    \label{fig:surrogate_dynamics}
\end{figure}

To further validate that GRAFT's soft-Spearman loss is a reliable proxy for the target evaluation metric, we track both the training loss and the test C-index across all epochs for each of the runs of GRAFT in Table \ref{tab:main_results}(3 seeds $\times$ 3 folds) per dataset. Figure~\ref{fig:surrogate_dynamics} shows the mean $\pm$ standard deviation of both curves on all six benchmarks. In every case, the loss decreases while the test C-index increases in tandem, confirming that minimizing the differentiable ranking objective consistently improves concordance. The tight standard deviation bands across runs further demonstrate stable convergence. The one notable exception is AIDS (92\% censoring), where the loss and C-index curves exhibit higher variance. This is consistent with the quantitative results in Table~\ref{tab:main_results}: with only $\sim$92 observed events, the local KM neighborhoods lack sufficient event information to produce stable imputation targets, weakening the gradient signal and limiting the loss's ability to reliably guide the model toward better concordance. This reinforces our recommendation that practitioners consider simpler parametric alternatives when event rates fall below 10\%.

\subsection{Ablation Study}
\label{sec:ablation_study}

\begin{figure}[t]
\centering
\includegraphics[width=0.7\textwidth,height=5cm]{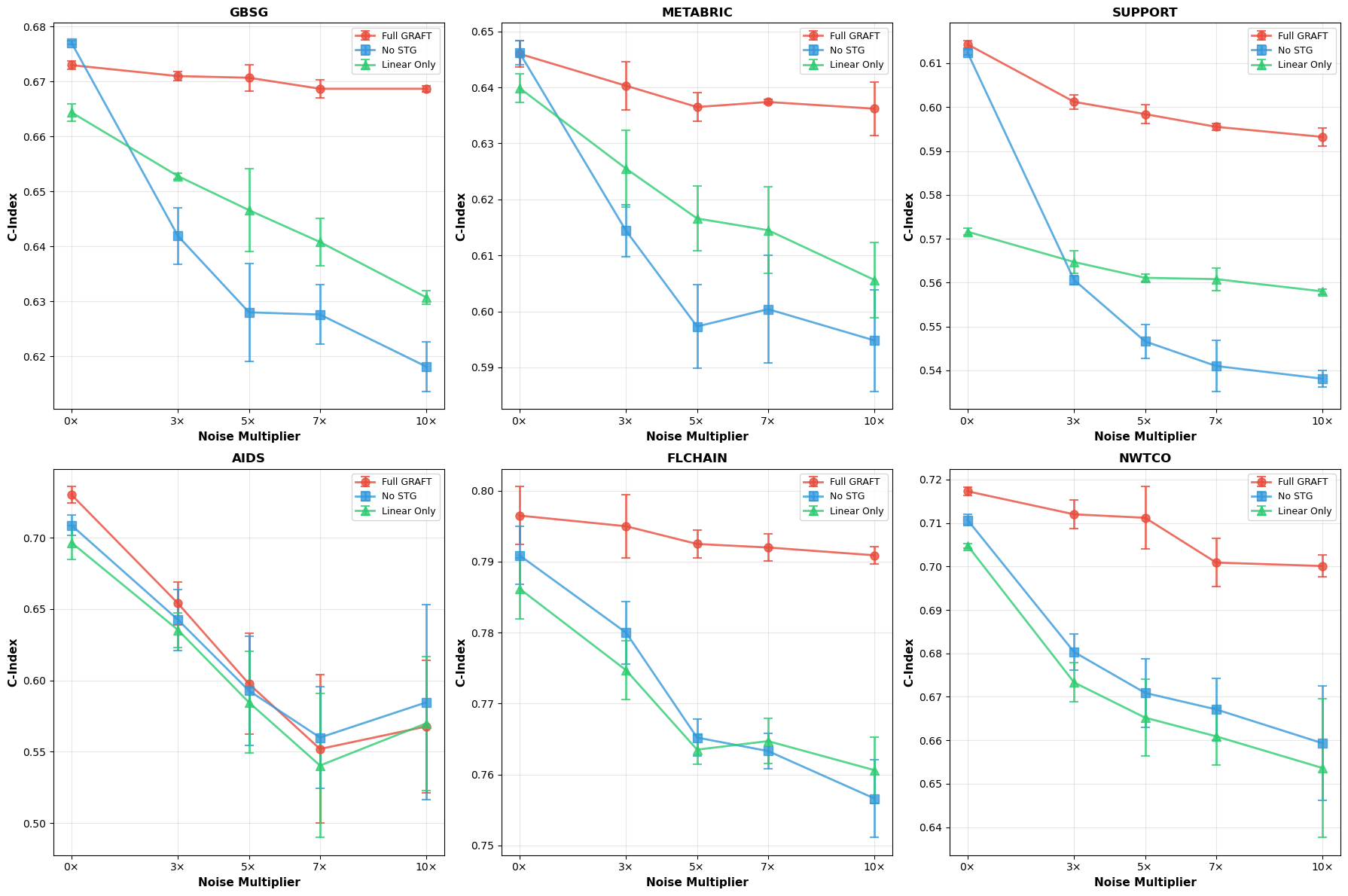}
\caption{\small{Ablation study showing C-index degradation under increasing Gaussian noise for three GRAFT variants. Error bars represent standard deviation across 3 random seeds.}}
\label{fig:ablation}
\end{figure}
Figure~\ref{fig:ablation} shows the C-index metrics across noise levels for the three GRAFT variants in all six datasets. The results reveal a clear performance hierarchy that validates our architectural design choices and quantifies the contribution of each component. At baseline (0x noise), {\em Full GRAFT} and {\em No STG} perform similarly on GBSG with a C-index of 0.673 and 0.676, respectively, while {\em Linear Only} achieves a lower C-index of 0.665. As noise increases, performance gaps widen dramatically. At 10x noise, {\em Full GRAFT} maintains a C-index of 0.669 (0.6\% drop from baseline), while {\em No STG} drops to 0.618 (8.5\% drop) and \enquote{Linear Only} drops to 0.631 (5.1\% drop). Similar patterns emerge on METABRIC, SUPPORT, FLCHAIN, and NWTCO.

Interestingly, {\em No STG} often performs worse than {\em Linear Only} at high noise levels, suggesting that the flexibility of the MLP becomes harmful when operating on noisy features without protective filtering. This interaction effect underscores GRAFT's integrated design: gates and MLP work synergistically, with gates enabling the MLP to focus on truly informative features rather than fitting to noise. Dataset-specific patterns are also informative. AIDS shows the smallest performance gaps, suggesting that high-censoring limits the benefit of complex modeling.
The calibration results shown in Appendix~\ref{sec:ibs_results} indicate that IBS follows a similar pattern, demonstrating that feature selection improves both discrimination and calibration.

\begin{figure}[t]
\centering
\includegraphics[width=0.7\textwidth,height=5cm]{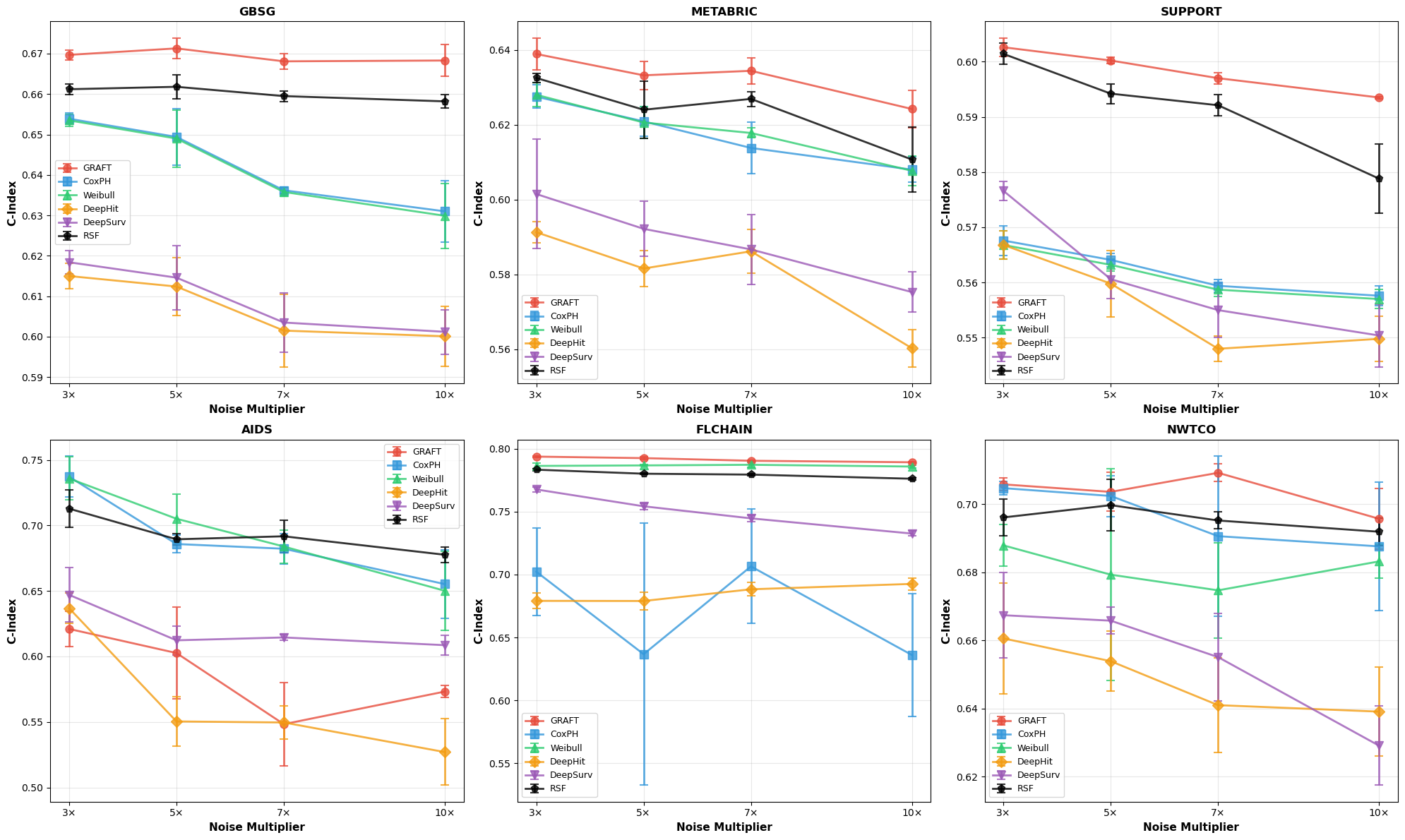}

\caption{\small{Comparison of C-index for all six models under heavy-tailed Student's $t$ noise ($df=2$). Error bars represent standard deviation across 3 random seeds.}}
\label{fig:noise_robustness}
\end{figure}

\subsection{Noise Robustness Analysis}
\label{sec:noise_robustness_results}

Figure~\ref{fig:noise_robustness} presents the C-index metrics for all six models under heavy-tailed Student's t noise ($df=2$) across increasing noise multipliers. This experiment reveals fundamental differences in how each model handles spurious high-dimensional features and provides direct evidence of GRAFT's superior robustness. On GBSG, GRAFT's C-index decreases minimally from 0.670 at 3x noise to 0.669 at 10x noise (0.15\% degradation), while DeepSurv drops from 0.619 to 0.601 (2.9\% degradation). On METABRIC, SUPPORT, FLCHAIN and NWTCO, GRAFT remains relatively stable across all noise levels. RSF emerges as the second best model in terms of robustness across different noise levels, with bootstrap sampling likely providing some implicit noise resistance. However, RSF's performance 
still degrades more than GRAFT. On SUPPORT, RSF's C-index drops from 0.602 at 3x noise to 0.579 at 
10x noise (3.8\% degradation), while GRAFT only drops from 0.604 to 0.594 (1.6\% degradation). The contrast between approaches is instructive. Classical models (CoxPH, Weibull) show moderate robustness but lack explicit feature selection, leading to gradual degradation as noise accumulates. RSF shows better robustness, though inferior to GRAFT's explicit feature gating. Deep learning models exhibit severe vulnerabilities: DeepSurv and DeepHit show catastrophic degradation, which is likely due to overfitting to noise patterns without mechanisms to distinguish signal from spurious features.

The key insight is that GRAFT's stochastic gates provide explicit, learnable noise filtering superior to implicit regularization. GRAFT learns a sparse binary mask that selectively suppresses uninformative features through differentiable L0 regularization. The nearly flat performance curves demonstrate that GRAFT successfully identifies true signal features and ignores injected noise, even when noise features outnumber signal features 10 to 1. Dataset-specific patterns reveal informative variations: FLCHAIN shows remarkable stability across multiple models, suggesting strong signal quality despite high censoring.
Moreover, GBSG, METABRIC, and SUPPORT show clear model separation with GRAFT's advantages most pronounced; while AIDS reinforces that extreme censoring (92\%) favors simpler parametric assumptions. The calibration results shown in Appendix \ref{sec:ibs_results} confirm that IBS follows similar trends, with GRAFT maintaining stable calibration performance, while competing models show increasing calibration errors under noise.

\subsection{Imputation Neighborhood Sensitivity Analysis}

\begin{figure}[t]
    \centering
    \includegraphics[width=0.9\textwidth]{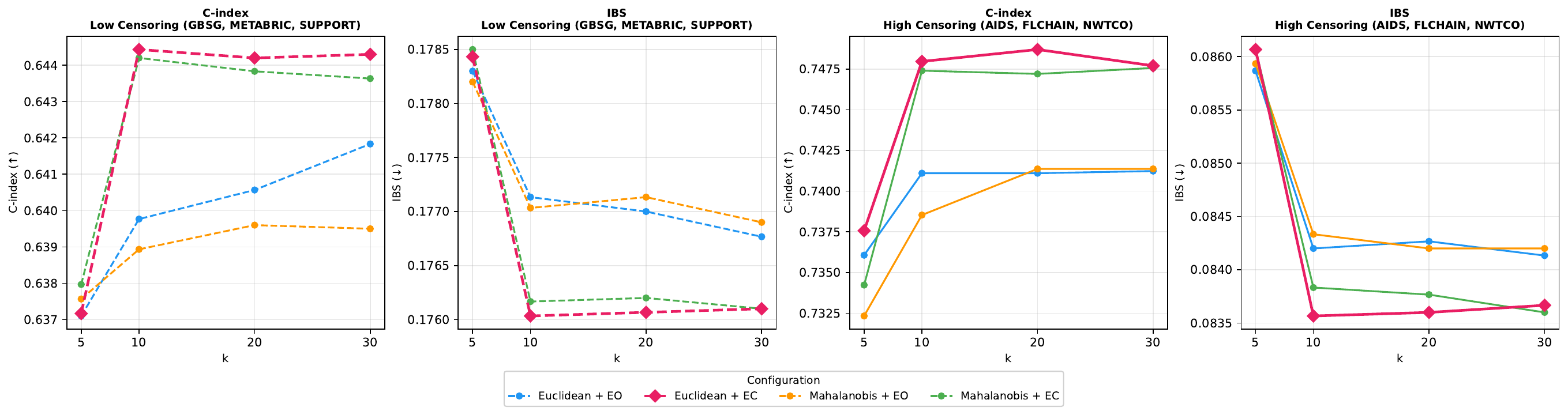}
    \caption{\small{Sensitivity of C-index (left) and IBS (right) to neighborhood
    construction choices across all six datasets. Each point is the mean across all runs.}}
    \label{fig:sensitivity}
\end{figure}

Figure~\ref{fig:sensitivity} reports the C-index and IBS for all neighborhood
construction configurations. Two findings emerge consistently in all six
datasets. First, performance stabilizes rapidly once $k$
reaches 10, with negligible changes at 20 and 30. This validates our choice of $k$ in the main experiments: requiring fewer events (5)
yields unstable KM estimates with marginally worse discrimination,
while requiring more (20, 30) offers no measurable benefit. Second, the distance metric and the neighborhood composition have minimal impact on performance. Both Euclidean
and Mahalanobis distances produce near-identical results across all configurations with Euclidean slightly outperforming Mahalanobis. EC consistently outperforms EO, indicating that including censored neighbors in the KM fit is essential to avoid biased estimates. These results demonstrate that GRAFT's imputation mechanism is robust to neighborhood construction choices. Detailed results are presented in Appendix~\ref{sc:full_imputation_results}.

\section{Conclusions}
\label{sec:conclusion}
We introduced GRAFT, a novel survival analysis model built around three core contributions: the explicit decoupling of prognostic ranking from survival calibration, a hybrid architecture that combines linear and non-linear components trained via stochastic imputation and a C-index-aligned ranking loss, and integrated feature selection via stochastic gates. Our experiments demonstrate that GRAFT outperforms baselines in both ranking and calibration. The ablation study further confirms that the architectural components contribute synergistically, with stochastic gates providing robust performance preservation under extreme noise conditions.

{\bf Limitations} GRAFT is particularly well-suited for high-dimensional biomedical applications where automated feature selection is critical. However, the model shows limitations under extreme censoring ($ \ge 92\%$), where simpler parametric models may be preferable due to insufficient event information for reliable local neighborhood estimation. Future work could explore adaptive architectures for extreme censoring regimes and extensions to competing risk scenarios.

\bibliography{main}

@article{kaplan1958,
  title={Nonparametric estimation from incomplete observations},
  author={Kaplan, Edward L and Meier, Paul},
  journal={Journal of the American Statistical Association},
  volume={53},
  number={282},
  pages={457--481},
  year={1958},
  publisher={Taylor \& Francis}
}

@article{cox1972,
  title={Regression models and life-tables},
  author={Cox, David R},
  journal={Journal of the Royal Statistical Society: Series B (Methodological)},
  volume={34},
  number={2},
  pages={187--220},
  year={1972},
  publisher={Wiley Online Library}
}

@article{ishwaran2008,
  title={Random survival forests},
  author={Ishwaran, Hemant and Kogalur, Udaya B and Blackstone, Eugene H and Lauer, Michael S},
  journal={The Annals of Applied Statistics},
  volume={2},
  number={3},
  pages={841--860},
  year={2008},
  publisher={Institute of Mathematical Statistics}
}

@article{katzman2018,
  title={DeepSurv: personalized treatment recommender system using a Cox proportional hazards deep neural network},
  author={Katzman, Jared L and Shaham, Uri and Cloninger, Alexander and Bates, Jonathan and Jiang, Tingting and Kluger, Yuval},
  journal={BMC Medical Research Methodology},
  volume={18},
  number={1},
  pages={1--12},
  year={2018},
  publisher={Springer}
}

@article{lee2018,
  title={DeepHit: A deep learning approach to survival analysis with competing risks},
  author={Lee, Changhee and Zame, William R and Yoon, Jinsung and van der Schaar, Mihaela},
  journal={Proceedings of the AAAI Conference on Artificial Intelligence},
  volume={32},
  number={1},
  year={2018}
}

@article{yamada2020,
  title={Feature selection using stochastic gates},
  author={Yamada, Yutaro and Lindenbaum, Ofir and Negahban, Sahand and Kluger, Yuval},
  journal={Proceedings of the 37th International Conference on Machine Learning},
  pages={10648--10659},
  year={2020},
  organization={PMLR}
}

@book{rubin1987,
  title={Multiple Imputation for Nonresponse in Surveys},
  author={Rubin, Donald B},
  year={1987},
  publisher={John Wiley \& Sons},
  address={New York}
}

@techreport{beran1981,
  title={Nonparametric regression with randomly censored survival data},
  author={Beran, Rudolf},
  year={1981},
  institution={University of California, Berkeley}
}

@article{dabrowska1989,
  title={Uniform consistency of the kernel conditional {K}aplan-{M}eier estimate},
  author={Dabrowska, Dorota M},
  journal={Annals of Statistics},
  volume={17},
  number={3},
  pages={1157--1167},
  year={1989}
}

@article{harrell1982,
  title={Evaluating the yield of medical tests},
  author={Harrell, Frank E and Califf, Robert M and Pryor, David B and Lee, Kerry L and Rosati, Robert A},
  journal={JAMA},
  volume={247},
  number={18},
  pages={2543--2546},
  year={1982}
}

@inproceedings{blondel2020,
  title={Fast differentiable sorting and ranking},
  author={Blondel, Mathieu and Teboul, Olivier and Berthet, Quentin and Djolonga, Josip},
  booktitle={International Conference on Machine Learning},
  pages={950--959},
  year={2020},
  organization={PMLR}
}

@article{kingma2014adam,
  title={Adam: A method for stochastic optimization},
  author={Kingma, Diederik P and Ba, Jimmy},
  journal={arXiv preprint arXiv:1412.6980},
  year={2014}
}

@article{clark2003,
  title={Survival analysis part I: basic concepts and first analyses},
  author={Clark, TG and Bradburn, MJ and Love, SB and Altman, DG},
  journal={British journal of cancer},
  volume={89},
  number={2},
  pages={232--238},
  year={2003},
  publisher={Nature Publishing Group}
}

@article{he2023review,
  title={Review for Dynamic Prediction in Clinical Survival Analysis},
  author={He, Weiyi},
  journal={arXiv preprint arXiv:2311.15743},
  year={2023}
}

@article{abbasi2024survival,
  title={Survival prediction landscape: an in-depth systematic literature review on activities, methods, tools, diseases, and databases},
  author={Abbasi, Ahtisham Fazeel and Asim, Muhammad Nabeel and Ahmed, Sheraz and Vollmer, Sebastian and Dengel, Andreas},
  journal={Frontiers in Artificial Intelligence},
  volume={7},
  pages={1428501},
  year={2024},
  publisher={Frontiers Media SA}
}

@book{meeker2014statistical,
  title={Statistical methods for reliability data},
  author={Meeker, William Q and Escobar, Luis A and Pascual, Francis G},
  year={2014},
  publisher={John Wiley \& Sons}
}

@article{dirick2017time,
  title={Time to default in credit scoring using survival analysis: a benchmark study},
  author={Dirick, Lore and Claeskens, Gerda and Baesens, Bart},
  journal={Journal of the Operational Research Society},
  volume={68},
  number={6},
  pages={652--665},
  year={2017},
  publisher={Taylor \& Francis}
}

@book{kalbfleisch2002statistical,
  title={The statistical analysis of failure time data},
  author={Kalbfleisch, John D and Prentice, Ross L},
  year={2002},
  publisher={John Wiley \& Sons}
}

@article{tibshirani1997,
  title={The lasso method for variable selection in the Cox model},
  author={Tibshirani, Robert},
  journal={Statistics in medicine},
  volume={16},
  number={4},
  pages={385--395},
  year={1997},
  publisher={Wiley Online Library}
}

@article{zou2005,
  title={Regularization and variable selection via the elastic net},
  author={Zou, Hui and Hastie, Trevor},
  journal={Journal of the Royal Statistical Society: Series B (Statistical Methodology)},
  volume={67},
  number={2},
  pages={301--320},
  year={2005},
  publisher={Wiley Online Library}
}

@article{zhao2006,
  title={On model selection consistency of Lasso},
  author={Zhao, Peng and Yu, Bin},
  journal={Journal of Machine learning research},
  volume={7},
  number={11},
  year={2006}
}

@article{saikia2017review,
  title={A review on accelerated failure time models},
  author={Saikia, Rinku and Barman, Manash Pratim},
  journal={International Journal of Statistics and Systems},
  volume={12},
  number={2},
  pages={311--322},
  year={2017}
}

@article{kassani2015survival,
  title={Survival analysis of drug abuse relapse in addiction treatment centers},
  author={Kassani, Aziz and Niazi, Mohsen and Hassanzadeh, Jafar and Menati, Rostam},
  journal={International journal of high risk behaviors \& addiction},
  volume={4},
  number={3},
  pages={e23402},
  year={2015}
}

@article{knaus1995support,
  title={The SUPPORT prognostic model: Objective estimates of survival for seriously ill hospitalized adults},
  author={Knaus, William A and Harrell, Frank E and Lynn, Joanne and Goldman, Lee and Phillips, Russell S and Connors, Alfred F and Dawson, Neal V and Fulkerson, William J and Califf, Robert M and Desbiens, Norman and others},
  journal={Annals of internal medicine},
  volume={122},
  number={3},
  pages={191--203},
  year={1995},
  publisher={American College of Physicians}
}

@article{curtis2012genomic,
  title={The genomic and transcriptomic architecture of 2,000 breast tumours reveals novel subgroups},
  author={Curtis, Christina and Shah, Sohrab P and Chin, Suet-Feung and Turashvili, Gulisa and Rueda, Oscar M and Dunning, Mark J and Speed, Doug and Lynch, Andy G and Samarajiwa, Shamith and Yuan, Yinyin and others},
  journal={Nature},
  volume={486},
  number={7403},
  pages={346--352},
  year={2012},
  publisher={Nature Publishing Group UK London}
}

@article{schumacher1994randomized,
  title={Randomized 2 x 2 trial evaluating hormonal treatment and the duration of chemotherapy in node-positive breast cancer patients. German Breast Cancer Study Group.},
  author={Schumacher, Martin and Bastert, G and Bojar, H and H{\"u}bner, K and Olschewski, M and Sauerbrei, W and Schmoor, C and Beyerle, C and Neumann, RL and Rauschecker, HF},
  journal={Journal of Clinical Oncology},
  volume={12},
  number={10},
  pages={2086--2093},
  year={1994}
}

@inproceedings{dispenzieri2012use,
  title={Use of nonclonal serum immunoglobulin free light chains to predict overall survival in the general population},
  author={Dispenzieri, Angela and Katzmann, Jerry A and Kyle, Robert A and Larson, Dirk R and Therneau, Terry M and Colby, Colin L and Clark, Raynell J and Mead, Graham P and Kumar, Shaji and Melton III, L Joseph and others},
  booktitle={Mayo Clinic Proceedings},
  volume={87},
  number={6},
  pages={517--523},
  year={2012},
  organization={Elsevier}
}

@article{hammer1997controlled,
  title={A controlled trial of two nucleoside analogues plus indinavir in persons with human immunodeficiency virus infection and CD4 cell counts of 200 per cubic millimeter or less},
  author={Hammer, Scott M and Squires, Kathleen E and Hughes, Michael D and Grimes, Janet M and Demeter, Lisa M and Currier, Judith S and Eron Jr, Joseph J and Feinberg, Judith E and Balfour Jr, Henry H and Deyton, Lawrence R and others},
  journal={New England Journal of Medicine},
  volume={337},
  number={11},
  pages={725--733},
  year={1997},
  publisher={Mass Medical Soc}
}

@article{breslow1999design,
  title={Design and analysis of two-phase studies with binary outcome applied to Wilms tumour prognosis},
  author={Breslow, Norman E and Chatterjee, Nilanjan},
  journal={Journal of the Royal Statistical Society: Series C (Applied Statistics)},
  volume={48},
  number={4},
  pages={457--468},
  year={1999},
  publisher={Wiley Online Library}
}

@article{wei1992accelerated,
  title={The accelerated failure time model: a useful alternative to the Cox regression model in survival analysis},
  author={Wei, Lee-Jen},
  journal={Statistics in medicine},
  volume={11},
  number={14-15},
  pages={1871--1879},
  year={1992},
  publisher={Wiley Online Library}
}

@article{williams1992simple,
  title={Simple statistical gradient-following algorithms for connectionist reinforcement learning},
  author={Williams, Ronald J},
  journal={Machine learning},
  volume={8},
  number={3},
  pages={229--256},
  year={1992},
  publisher={Springer}
}

@article{graf1999assessment,
  title={Assessment and comparison of prognostic classification schemes for survival data},
  author={Graf, Erika and Schmoor, Claudia and Sauerbrei, Willi and Schumacher, Martin},
  journal={Statistics in medicine},
  volume={18},
  number={17-18},
  pages={2529--2545},
  year={1999},
  publisher={Wiley Online Library}
}

@article{devroye2006nonuniform,
  title={Nonuniform random variate generation},
  author={Devroye, Luc},
  journal={Handbooks in operations research and management science},
  volume={13},
  pages={83--121},
  year={2006},
  publisher={Elsevier}
}

@article{dabrowska1987non,
  title={Non-parametric regression with censored survival time data},
  author={Dabrowska, Dorota M},
  journal={Scandinavian Journal of Statistics},
  pages={181--197},
  year={1987},
  publisher={JSTOR}
}

@article{lin2007breslow,
  title={On the Breslow estimator},
  author={Lin, DY},
  journal={Lifetime data analysis},
  volume={13},
  number={4},
  pages={471--480},
  year={2007},
  publisher={Springer}
}

@article{miller2017reducing,
  title={Reducing reparameterization gradient variance},
  author={Miller, Andrew and Foti, Nick and D'Amour, Alexander and Adams, Ryan P},
  journal={Advances in Neural Information Processing Systems},
  volume={30},
  year={2017}
}

@article{best2000minimizing,
  title={Minimizing separable convex functions subject to simple chain constraints},
  author={Best, Michael J and Chakravarti, Nilotpal and Ubhaya, Vasant A},
  journal={SIAM Journal on Optimization},
  volume={10},
  number={3},
  pages={658--672},
  year={2000},
  publisher={SIAM}
}

@article{tibshirani2011nearly,
  title={Nearly-isotonic regression},
  author={Tibshirani, Ryan J and Hoefling, Holger and Tibshirani, Robert},
  journal={Technometrics},
  volume={53},
  number={1},
  pages={54--61},
  year={2011},
  publisher={Taylor \& Francis}
}

@article{burk2024large,
  title={A Large-Scale Neutral Comparison Study of Survival Models on Low-Dimensional Data},
  author={Burk, Lukas and Zobolas, John and Bischl, Bernd and Bender, Andreas and Wright, Marvin N and Sonabend, Raphael},
  journal={arXiv preprint arXiv:2406.04098},
  year={2024}
}

\newpage
\appendix
\onecolumn
\section{Broader Impact}

This paper presents work whose goal is to advance the field of machine learning in healthcare, specifically survival analysis for time-to-event prediction. Our work has direct applications in healthcare, where survival models inform critical clinical decisions including treatment planning, risk stratification, and resource allocation. While improved prediction accuracy can lead to better patient outcomes, we acknowledge several important considerations.

First, survival models trained on historical data may perpetuate existing biases in healthcare delivery if demographic disparities are present in training data. Practitioners should carefully audit model predictions across patient subgroups before clinical deployment. Second, the interpretability provided by our linear AFT component and feature selection mechanism, while beneficial for model understanding, should not substitute for clinical judgment—models should augment rather than replace physician decision-making. Third, the automatic feature selection capability, though valuable for identifying relevant biomarkers, requires validation that selected features reflect true biological mechanisms rather than spurious correlations.

We emphasize that GRAFT, like all machine learning models for healthcare, requires rigorous clinical validation, fairness auditing, and regulatory approval before deployment in medical settings. The model's performance limitations under extreme censoring ($\ge 92\%$) highlight the importance of understanding applicability boundaries. Responsible use requires ongoing monitoring, transparent communication of uncertainty, and integration within broader clinical workflows that maintain human oversight.
\section{Hyperparameters}
\label{sec:hyperparams}

For reproducibility purposes, all hyperparameters used for GRAFT and baseline models in the experiments are specified below in Table \ref{tab:hyperparameters}.

\vspace{-0.5cm}

\begin{table}[h]
\centering
\caption{Hyperparameters for all models used in experiments. All deep learning models use comparable hyperparameters to ensure fair comparison.}
\label{tab:hyperparameters}
\small
\begin{tabular}{@{}llp{6cm}@{}}
\toprule
\textbf{Model} & \textbf{Hyperparameter} & \textbf{Value} \\
\midrule
\multirow{14}{*}{\textbf{GRAFT}} 
& Hidden dimension ($d_h$) & 32 \\
& Learning rate & $10^{-3}$ \\
& Weight decay ($\alpha_{\text{L2}}$) & $10^{-4}$ \\
& L0 regularization ($\lambda_{\text{L0}}$) & 0.01 \\
& Dropout rate & 0.2 \\
& STG noise std ($\sigma$) & 0.5 \\
& Soft-rank smoothing ($\tau$) & 0.1 \\
& Batch size ($m$) & 64 \\
& Monte Carlo samples ($M$) & 5 \\
& Training epochs ($E$) & 1000 with early stopping \\
& Patience & 10 \\
& Min obs events per neighborhood (k) & 10 \\
& Distance metric & Euclidean \\
& Optimizer & Adam \\
\midrule
\multirow{1}{*}{\textbf{CoxPH}} 
& L2 penalty (penalizer) & 0.01 \\
\midrule
\multirow{1}{*}{\textbf{Weibull AFT}} 
& L2 penalty (penalizer) & 0.01 \\
\midrule
\multirow{11}{*}{\textbf{DeepHit}} 
& Number of duration bins & 100 \\
& Hidden layers & [32, 32] \\
& Batch normalization & True \\
& Dropout rate & 0.2 \\
& Learning rate & 0.01 \\
& Optimizer & Adam \\
& Alpha (ranking loss weight) & 0.3 \\
& Sigma (smoothing parameter) & 0.1 \\
& Batch size & 64 \\
& Training epochs & 1000 with early stopping \\
& Patience & 10 \\
\midrule
\multirow{9}{*}{\textbf{DeepSurv}} 
& Hidden layers & [32, 32] \\
& Batch normalization & True \\
& Dropout rate & 0.2 \\
& Learning rate & 0.01 \\
& Optimizer & Adam \\
& Output bias & False \\
& Batch size & 64 \\
& Training epochs & 1000 with early stopping \\
& Patience & 10 \\
\midrule
\multirow{5}{*}{\textbf{RSF}} 
& Number of trees & 100 \\
& Min samples split & 10 \\
& Min samples leaf & 15 \\
& Max features & sqrt \\
\bottomrule
\end{tabular}
\end{table}

\clearpage

\section{Additional Results and Experiments}
\label{sec:additional_exp}

\subsection{IBS Results for Ablation and Noise Experiment}
\label{sec:ibs_results}
\begin{figure*}[t]
\centering
\includegraphics[width=0.70\textwidth]{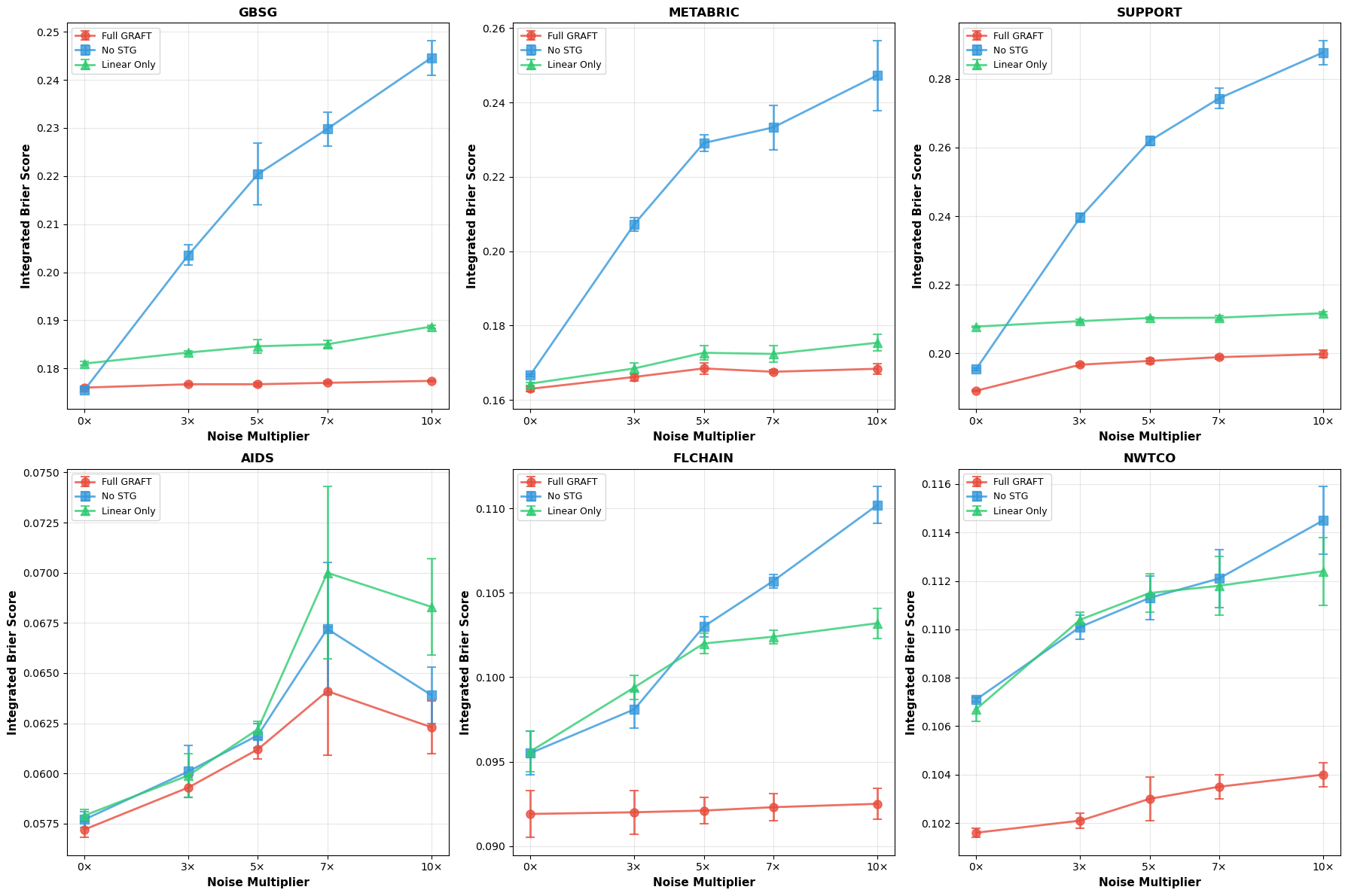}
\caption{\small{Ablation study showing IBS degradation under increasing Gaussian noise for three GRAFT variants across six datasets. \enquote{Full GRAFT} (red) maintains stable performance across noise levels. Error bars represent standard deviation across 3 random seeds.}}
\label{fig:ablation_ibs}
\end{figure*}

\begin{figure*}[t]
\centering
\includegraphics[width=0.70\textwidth]{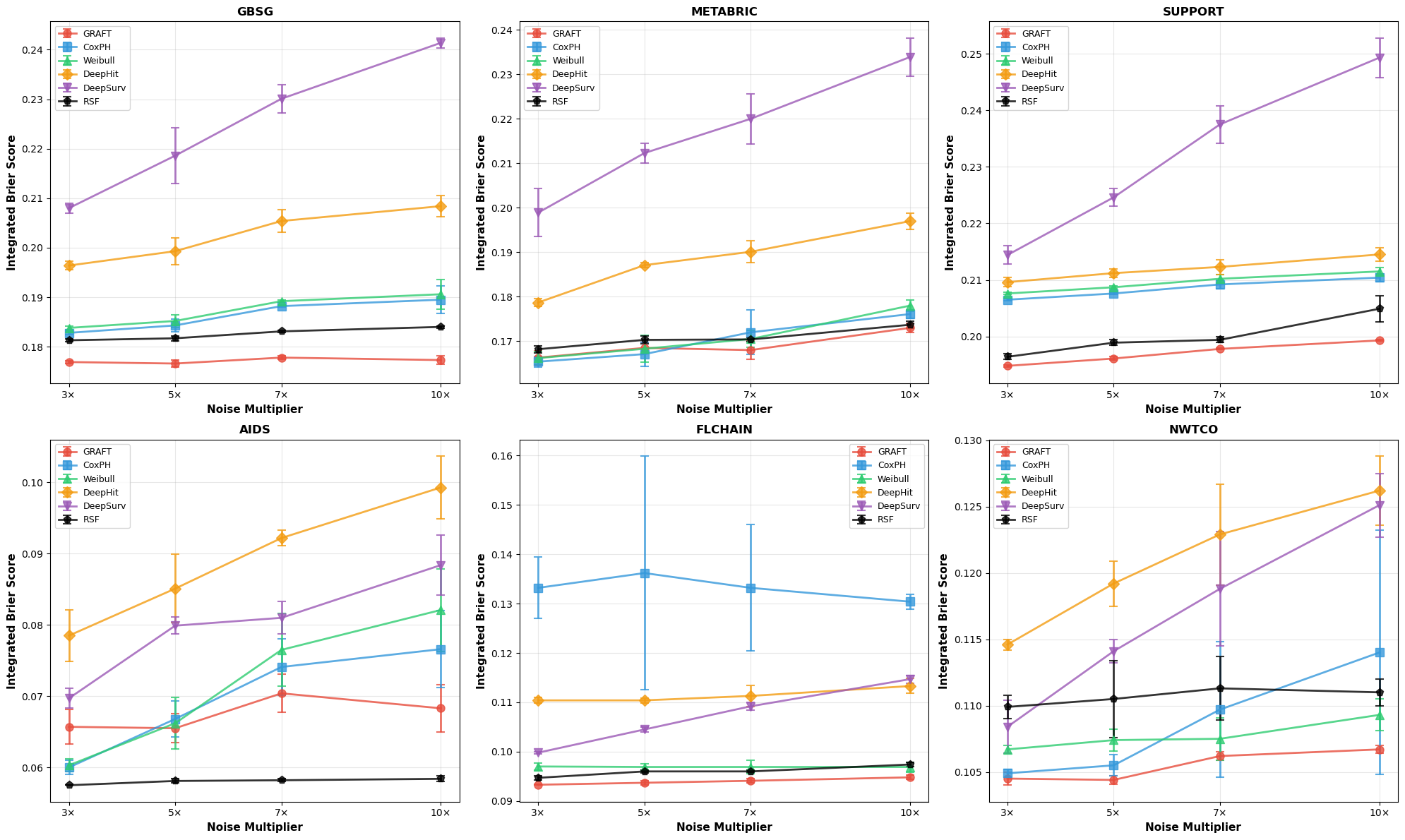}
\caption{\small{Noise robustness comparison showing IBS under increasing Student's t noise ($df=2$) for all six models across six datasets. GRAFT (red) maintains the flattest performance curves, demonstrating superior robustness to heavy-tailed noise. Error bars represent standard deviation across 3 random seeds.}}
\label{fig:noise_robustness_ibs}
\end{figure*}

The ablation study (Sec \ref{sec:ablation_study}) and noise robustness analysis (Sec \ref{sec:noise_robustness_results}) presented C-index results. Figures~\ref{fig:ablation_ibs} and~\ref{fig:noise_robustness_ibs} present the corresponding IBS results across all six datasets. The calibration metrics follow the same qualitative pattern. In Figure \ref{fig:ablation_ibs} {\em Full GRAFT} maintains stable IBS as noise increases, while the {\em No STG} variant shows degradation due to overfitting on irrelevant features, and the {\em Linear Only} variant exhibits limited flexibility. Figure \ref{fig:noise_robustness_ibs}  demonstrates GRAFT's superior discrimination performance under increasing noise. These results confirm that GRAFT's stochastic gates improve both discrimination and calibration in high-dimensional, noisy settings.

\subsection{Full Results for Imputation Neighbourhood Sensitivity Analysis}
\label{sc:full_imputation_results}
Tables~\ref{tab:sens_cindex} and~\ref{tab:sens_ibs} report the complete C-index and IBS results for all neighbourhood construction configurations across all six datasets. Each cell reports the mean and standard deviation across all runs (3 seeds $\times$ 3 folds). The configurations vary along three dimensions: the minimum number of observed events required within a neighbourhood (\texttt{min\_events} $\in \{5, 10, 20, 30\}$), the distance metric (Euclidean or Mahalanobis), and the neighbourhood composition, where EO (Events-Only) restricts the local Kaplan-Meier estimator to neighbours with observed events, while EC (Events+Censored) includes all neighbours regardless of censoring status. All other hyperparameters are fixed at the values used in the main experiments. The configuration used in the paper (\texttt{k=10}, Euclidean, EC) is marked with $\dagger$.

\begin{table*}[t]
\centering
\caption{\small{Harrell's C-index (mean $\pm$ std across all runs, higher is better) for all
neighbourhood construction configurations across six datasets.
Low-censoring: GBSG (43\%), METABRIC (42\%), SUPPORT (32\%).
High-censoring: AIDS (92\%), FLCHAIN (72\%), NWTCO (86\%).
Best value per dataset in \textbf{bold}. $\dagger$~Paper configuration.
EO = Events-Only; EC = Events+Censored.}}
\label{tab:sens_cindex}
\begin{adjustbox}{max width=\textwidth}
\small
\begin{tabular}{@{}ccccccccc@{}}
\toprule
\textbf{k} & \textbf{Dist.} & \textbf{Comp.} & \textbf{GBSG} & \textbf{METABRIC} & \textbf{SUPPORT} & \textbf{AIDS} & \textbf{FLCHAIN} & \textbf{NWTCO} \\
\midrule
\multicolumn{9}{l}{\textit{Low censoring datasets: GBSG, METABRIC, SUPPORT — High censoring datasets: AIDS, FLCHAIN, NWTCO}} \\
\midrule
5 & Euc & EO & 0.6649 (0.0021) & 0.6366 (0.0037) & 0.6098 (0.0011) & 0.7105 (0.0052) & 0.7859 (0.0025) & 0.7118 (0.0014) \\
5 & Euc & EC & 0.6662 (0.0011) & 0.6371 (0.0038) & 0.6082 (0.0019) & 0.7111 (0.0069) & 0.7889 (0.0032) & 0.7127 (0.0017) \\
5 & Mah & EO & 0.6662 (0.0024) & 0.6368 (0.0028) & 0.6097 (0.0013) & 0.6960 (0.0071) & 0.7887 (0.0024) & 0.7123 (0.0018) \\
5 & Mah & EC & 0.6669 (0.0019) & 0.6378 (0.0042) & 0.6092 (0.0016) & 0.7037 (0.0064) & 0.7860 (0.0027) & 0.7130 (0.0016) \\
\midrule
10 & Euc & EO & 0.6693 (0.0015) & 0.6390 (0.0015) & 0.6110 (0.0013) & 0.7173 (0.0045) & 0.7897 (0.0010) & 0.7163 (0.0017) \\
10 & Euc & EC\,$\dagger$ & \textbf{0.6730} (0.0008) & 0.6460 (0.0016) & \textbf{0.6143} (0.0010) & 0.7301 (0.0039) & 0.7965 (0.0004) & 0.7173 (0.0012) \\
10 & Mah & EO & 0.6690 (0.0008) & 0.6385 (0.0014) & 0.6093 (0.0012) & 0.7113 (0.0060) & 0.7884 (0.0009) & 0.7159 (0.0013) \\
10 & Mah & EC & 0.6728 (0.0010) & \textbf{0.6461} (0.0017) & 0.6137 (0.0017) & 0.7292 (0.0047) & 0.7955 (0.0005) & 0.7175 (0.0015) \\
\midrule
20 & Euc & EO & 0.6694 (0.0012) & 0.6398 (0.0017) & 0.6125 (0.0061) & 0.7176 (0.0061) & 0.7898 (0.0014) & 0.7159 (0.0013) \\
20 & Euc & EC & 0.6728 (0.0010) & 0.6458 (0.0014) & 0.6140 (0.0009) & \textbf{0.7328} (0.0036) & \textbf{0.7966} (0.0007) & 0.7167 (0.0012) \\
20 & Mah & EO & 0.6692 (0.0011) & 0.6392 (0.0014) & 0.6104 (0.0046) & 0.7182 (0.0057) & 0.7899 (0.0013) & 0.7160 (0.0013) \\
20 & Mah & EC & 0.6718 (0.0009) & 0.6459 (0.0010) & 0.6138 (0.0018) & 0.7280 (0.0041) & 0.7963 (0.0004) & 0.7173 (0.0012) \\
\midrule
30 & Euc & EO & 0.6692 (0.0007) & 0.6428 (0.0017) & 0.6135 (0.0012) & 0.7181 (0.0032) & 0.7897 (0.0008) & 0.7159 (0.0011) \\
30 & Euc & EC & \textbf{0.6730} (0.0012) & 0.6458 (0.0016) & 0.6141 (0.0011) & 0.7295 (0.0047) & 0.7964 (0.0006) & 0.7172 (0.0015) \\
30 & Mah & EO & 0.6693 (0.0009) & 0.6394 (0.0010) & 0.6098 (0.0015) & 0.7189 (0.0035) & 0.7892 (0.0014) & 0.7160 (0.0010) \\
30 & Mah & EC & 0.6728 (0.0008) & 0.6451 (0.0012) & 0.6130 (0.0019) & 0.7289 (0.0053) & 0.7962 (0.0004) & \textbf{0.7176} (0.0014) \\
\bottomrule
\end{tabular}
\end{adjustbox}
\end{table*}

\begin{table*}[t]
\centering
\caption{\small{Integrated Brier Score (mean $\pm$ std across all runs, lower is better) for all
neighbourhood construction configurations across six datasets.
Calibration obtained via post-hoc 1D CoxPH on GRAFT scores, identical to
the main experiments. Best value per dataset in \textbf{bold}.
$\dagger$~Paper configuration. EO = Events-Only; EC = Events+Censored.}}
\label{tab:sens_ibs}
\begin{adjustbox}{max width=\textwidth}
\small
\begin{tabular}{@{}ccccccccc@{}}
\toprule
\textbf{k} & \textbf{Dist.} & \textbf{Comp.} & \textbf{GBSG} & \textbf{METABRIC} & \textbf{SUPPORT} & \textbf{AIDS} & \textbf{FLCHAIN} & \textbf{NWTCO} \\
\midrule
\multicolumn{9}{l}{\textit{Low censoring datasets: GBSG, METABRIC, SUPPORT — High censoring datasets: AIDS, FLCHAIN, NWTCO}} \\
\midrule
5 & Euc & EO & 0.1783 (0.0008) & 0.1632 (0.0013) & 0.1934 (0.0024) & 0.0591 (0.0017) & 0.0951 (0.0019) & 0.1034 (0.0014) \\
5 & Euc & EC & 0.1784 (0.0006) & 0.1634 (0.0017) & 0.1935 (0.0022) & 0.0591 (0.0016) & 0.0954 (0.0020) & 0.1037 (0.0011) \\
5 & Mah & EO & 0.1780 (0.0006) & 0.1632 (0.0020) & 0.1934 (0.0025) & 0.0592 (0.0015) & 0.0954 (0.0026) & 0.1032 (0.0013) \\
5 & Mah & EC & 0.1785 (0.0006) & 0.1633 (0.0014) & 0.1937 (0.0022) & 0.0592 (0.0017) & 0.0951 (0.0026) & 0.1039 (0.0012) \\
\midrule
10 & Euc & EO & 0.1764 (0.0006) & 0.1631 (0.0004) & 0.1919 (0.0008) & 0.0581 (0.0006) & 0.0922 (0.0002) & 0.1023 (0.0005) \\
10 & Euc & EC\,$\dagger$ & \textbf{0.1760} (0.0002) & 0.1630 (0.0007) & \textbf{0.1891} (0.0005) & 0.0572 (0.0004) & \textbf{0.0919} (0.0001) & \textbf{0.1016} (0.0003) \\
10 & Mah & EO & 0.1765 (0.0006) & 0.1631 (0.0006) & 0.1915 (0.0012) & 0.0583 (0.0007) & 0.0925 (0.0006) & 0.1022 (0.0004) \\
10 & Mah & EC & \textbf{0.1760} (0.0005) & 0.1628 (0.0007) & 0.1897 (0.0007) & 0.0575 (0.0006) & 0.0921 (0.0007) & 0.1019 (0.0005) \\
\midrule
20 & Euc & EO & 0.1761 (0.0009) & 0.1630 (0.0007) & 0.1919 (0.0005) & 0.0582 (0.0009) & 0.0923 (0.0008) & 0.1023 (0.0006) \\
20 & Euc & EC & 0.1761 (0.0009) & 0.1629 (0.0006) & 0.1892 (0.0004) & 0.0572 (0.0008) & \textbf{0.0919} (0.0004) & 0.1017 (0.0003) \\
20 & Mah & EO & 0.1762 (0.0007) & 0.1631 (0.0012) & 0.1921 (0.0009) & 0.0583 (0.0007) & 0.0923 (0.0005) & 0.1020 (0.0005) \\
20 & Mah & EC & 0.1762 (0.0004) & \textbf{0.1626} (0.0010) & 0.1898 (0.0006) & 0.0574 (0.0011) & 0.0920 (0.0004) & 0.1019 (0.0004) \\
\midrule
30 & Euc & EO & 0.1763 (0.0004) & 0.1629 (0.0006) & 0.1911 (0.0006) & 0.0581 (0.0007) & 0.0921 (0.0007) & 0.1022 (0.0006) \\
30 & Euc & EC & 0.1761 (0.0005) & 0.1629 (0.0008) & 0.1893 (0.0009) & \textbf{0.0571} (0.0008) & 0.0920 (0.0005) & 0.1019 (0.0007) \\
30 & Mah & EO & 0.1762 (0.0004) & 0.1630 (0.0005) & 0.1915 (0.0010) & 0.0583 (0.0006) & 0.0922 (0.0006) & 0.1021 (0.0005) \\
30 & Mah & EC & 0.1763 (0.0003) & 0.1628 (0.0011) & 0.1892 (0.0007) & 0.0573 (0.0007) & \textbf{0.0919} (0.0003) & \textbf{0.1016} (0.0004) \\
\bottomrule
\end{tabular}
\end{adjustbox}
\end{table*}

\clearpage

\subsection{Alternative Method for Survival Curve Estimation: Isotonic Regression}
\label{sec:isotonic_calibration}
In Section~\ref{sec:survival_estimation}, we employ Cox Proportional Hazards (CoxPH) to calibrate GRAFT's prognostic scores into survival curves. While this approach is computationally efficient and provides smooth, well-calibrated survival estimates, it introduces a potential methodological tension: Cox calibration step explicitly assumes that hazard ratios remain constant over time. To address this, we evaluate an alternative non-parametric calibration method based on isotonic regression. For a grid of $K=50$ discrete time points spanning the test set's time range, we fit one isotonic regression at each time point. At a given time point $t_k$ (for $k=1,\ldots,K$), we include subjects with known survival status at $t_k$: those with $t_i > t_k$ (who survived past $t_k$), and those with $\delta_i = 1$ and $t_i \leq t_k$ (whose event occurred by $t_k$). Subjects with $\delta_i = 0$ and $t_i \leq t_k$ are excluded, as their survival status at $t_k$ is ambiguous. We construct binary survival indicators as targets: $y_i^{(k)} = 1$ for subjects with $t_i > t_k$ and $y_i^{(k)} = 0$ for subjects with $\delta_i = 1$ and $t_i \leq t_k$. Isotonic regression then learns a function from GRAFT scores to these binary targets, yielding calibrated survival probability estimates $\hat{S}(t_k | s_i) \in [0,1]$ where $s_i$ denotes the GRAFT score for subject $i$. Survival curves are constructed by linear interpolation between the $K=50$ calibrated probability estimates.

This isotonic approach offers two potential advantages over Cox: (1) it makes no parametric assumptions about the baseline hazard or proportional hazards, and (2) each time point learns an independent score-to-probability mapping, allowing the relationship to vary over time (capturing non-proportional hazards effects if present). However, it also has potential drawbacks: (1) reduced sample size at each time point due to censoring, especially at later times, and (2) interpolation artifacts between discrete time points. Table~\ref{tab:isotonic_vs_cox} compares Integrated Brier Scores (IBS) for GRAFT calibrated with Cox vs isotonic regression across all six benchmark datasets. As shown in Table~\ref{tab:isotonic_vs_cox}, both calibration methods achieve competitive performance. Cox calibration demonstrates better mean IBS on five of six datasets and lower standard deviations across both fold-averaged and seed-averaged metrics. These properties—competitive calibration accuracy, lower variance, and computational efficiency motivated our choice of Cox calibration for GRAFT. However, the comparable performance of isotonic regression demonstrates that practitioners may opt for isotonic regression when non-proportional hazards effects are strongly suspected or when fully non-parametric calibration is desired.

\begin{table}[htbp]
\centering
\caption{Comparison of calibration methods: Cox PH vs Isotonic Regression. Lower IBS is better. Values reported as Mean (Fold-Std, Seed-Std).}
\label{tab:isotonic_vs_cox}
\begin{tabular}{lccc}
\toprule
\textbf{Dataset} & \textbf{GRAFT} & \textbf{GRAFT w/ Isotonic} &  \\
\midrule
GBSG       & \textbf{0.1760} (0.0002, 0.0012) & 0.1782 (0.0004, 0.0014) & \\
METABRIC   & 0.1630 (0.0007, 0.0045) & \textbf{0.1627} (0.0009, 0.0048) & \\
SUPPORT    & \textbf{0.1891} (0.0005, 0.0018) & 0.1902 (0.0006, 0.0024) & \\
NWTCO      & \textbf{0.1016} (0.0003, 0.0007) & 0.1032 (0.0006, 0.0010) & \\
FLCHAIN    & \textbf{0.0919} (0.0001, 0.0004) & 0.0927 (0.0003, 0.0007) & \\
AIDS       & \textbf{0.0572} (0.0004, 0.0032) & 0.0602 (0.0009, 0.0045) & \\

\bottomrule
\end{tabular}
\end{table}
\subsection{Gating Mechanism Ablation Study}

To validate our choice of Stochastic Gates (STG) as the feature selection mechanism in GRAFT, we conducted an ablation study comparing three gating approaches across all six benchmark datasets across. For a dataset with $p$ original features, we augment it with $kp$ additional noise features at multipliers $k \in \{5, 10\}$, where each noise feature is independently sampled from a heavy-tailed Student's t-distribution with 2 degrees of freedom. The three variants tested were: (1) \textbf{STG}: a differentiable feature selection mechanism based on continuous relaxation of Bernoulli variables \cite{yamada2020}; (2) \textbf{Sigmoid}: a simpler deterministic gating mechanism using learnable parameters $\alpha \in \mathbb{R}^p$, where gates are computed as $g_j = \text{sigmoid}(\alpha_j)$ for each j-th feature, without stochasticity; and (3) \textbf{REINFORCE}: a policy gradient approach \citep{williams1992simple} where binary gates are sampled from a Bernoulli distribution and updated using the REINFORCE algorithm with variance reduction and entropy regularization. All three variants use the same GRAFT architecture (linear AFT + residual MLP), the same local KM imputation procedure, and the same soft-ranking training objective—only the gating mechanism differs. Table~\ref{tab:gating_comparison} presents the results across all datasets. At 0x noise, all three gating mechanisms perform similarly; for example, in FLCHAIN, STG achieves a C-index of $0.7965$, while Sigmoid and REINFORCE achieve $0.7950$ and $0.7947$, respectively. However, as noisy features increase, STG consistently demonstrates superior performance preservation, maintaining better discrimination (C-Index) and calibration (IBS) compared to the alternatives. REINFORCE ranks second, showing moderate robustness to noise, while Sigmoid exhibits significant performance degradation under noisy conditions. 

\begin{table}[b]
\footnotesize
\setlength{\tabcolsep}{3pt}
\centering
\caption{Ablation study comparing gating mechanisms across six datasets under increasing noise conditions. Values reported as Mean (Fold-Std, Seed-Std) based on 3-fold cross-validation with 3 random seeds. Noise levels: 0x, 5x, 10x. Best values per dataset and noise level in bold.}
\label{tab:gating_comparison}
\begin{tabular}{llcccc}
\toprule
\textbf{Metric} & \textbf{Dataset} & \textbf{Noise} & \textbf{STG} & \textbf{Sigmoid} & \textbf{REINFORCE} \\
\midrule
\multicolumn{6}{l}{\textit{Low Censoring Datasets}} \\
\midrule
C-Index & GBSG & 0x & \textbf{0.6730} (0.0008, 0.0022) & 0.6691 (0.0020, 0.0035) & 0.6722 (0.0015, 0.0028) \\
(higher) & & 5x & \textbf{0.6713} (0.0026, 0.0026) & 0.6583 (0.0040, 0.0055) & 0.6680 (0.0030, 0.0042) \\
        & & 10x & 0.6683 (0.0019, 0.0039) & 0.6450 (0.0050, 0.0080) & \textbf{0.6690} (0.0038, 0.0065) \\
\cmidrule{2-6}
        & METABRIC & 0x & \textbf{0.6460} (0.0016, 0.0046) & 0.6421 (0.0030, 0.0065) & 0.6350 (0.0025, 0.0055) \\
        & & 5x & 0.6332 (0.0038, 0.0038) & \textbf{0.6334} (0.0060, 0.0075) & 0.6280 (0.0052, 0.0065) \\
        & & 10x & \textbf{0.6242} (0.0035, 0.0050) & 0.5990 (0.0075, 0.0095) & 0.6184 (0.0065, 0.0072) \\
\cmidrule{2-6}
        & SUPPORT & 0x & \textbf{0.6143} (0.0010, 0.0005) & 0.5992 (0.0025, 0.0020) & 0.6050 (0.0018, 0.0012) \\
        & & 5x & \textbf{0.6002} (0.0006, 0.0006) & 0.5786 (0.0020, 0.0030) & 0.5959 (0.0015, 0.0028) \\
        & & 10x & \textbf{0.5935} (0.0010, 0.0002) & 0.5681 (0.0035, 0.0040) & 0.5882 (0.0025, 0.0028) \\
\midrule
IBS & GBSG & 0x & \textbf{0.1760} (0.0002, 0.0012) & 0.1761 (0.0010, 0.0022) & \textbf{0.1760} (0.0008, 0.0018) \\
(lower) & & 5x & 0.1766 (0.0007, 0.0007) & 0.1920 (0.0018, 0.0030) & \textbf{0.1755} (0.0012, 0.0020) \\
        & & 10x & \textbf{0.1773} (0.0003, 0.0009) & 0.1972 (0.0025, 0.0045) & 0.1850 (0.0018, 0.0035) \\
\cmidrule{2-6}
        & METABRIC & 0x & \textbf{0.1630} (0.0007, 0.0045) & 0.1682 (0.0020, 0.0065) & 0.1655 (0.0015, 0.0058) \\
        & & 5x & \textbf{0.1685} (0.0013, 0.0013) & 0.1859 (0.0035, 0.0075) & 0.1750 (0.0028, 0.0065) \\
        & & 10x & \textbf{0.1730} (0.0020, 0.0011) & 0.2002 (0.0045, 0.0090) & 0.1824 (0.0038, 0.0075) \\
\cmidrule{2-6}
        & SUPPORT & 0x & \textbf{0.1891} (0.0005, 0.0018) & 0.1901 (0.0015, 0.0035) & 0.1899 (0.0012, 0.0028) \\
        & & 5x & \textbf{0.1961} (0.0003, 0.0003) & 0.1988 (0.0020, 0.0040) & 0.1974 (0.0015, 0.0032) \\
        & & 10x & \textbf{0.1993} (0.0001, 0.0001) & 0.2282 (0.0030, 0.0050) & 0.2080 (0.0020, 0.0040) \\
\midrule
\multicolumn{6}{l}{\textit{High Censoring Datasets}} \\
\midrule
C-Index & NWTCO & 0x & \textbf{0.7173} (0.0012, 0.0055) & 0.7122 (0.0035, 0.0075) & 0.7080 (0.0025, 0.0065) \\
(higher) & & 5x & 0.7036 (0.0057, 0.0057) & 0.6780 (0.0080, 0.0115) & \textbf{0.7051} (0.0070, 0.0095) \\
        & & 10x & \textbf{0.6957} (0.0026, 0.0089) & 0.6620 (0.0065, 0.0120) & 0.6921 (0.0050, 0.0100) \\
\cmidrule{2-6}
        & FLCHAIN & 0x & \textbf{0.7965} (0.0004, 0.0010) & 0.7950 (0.0018, 0.0030) & 0.7947 (0.0010, 0.0022) \\
        & & 5x & \textbf{0.7926} (0.0009, 0.0009) & 0.7683 (0.0025, 0.0040) & 0.7880 (0.0018, 0.0030) \\
        & & 10x & \textbf{0.7893} (0.0002, 0.0004) & 0.7591 (0.0035, 0.0055) & 0.7820 (0.0025, 0.0040) \\
\cmidrule{2-6}
        & AIDS & 0x & 0.7301 (0.0039, 0.0244) & 0.7296 (0.0180, 0.0300) & \textbf{0.7314} (0.0155, 0.0265) \\
        & & 5x & \textbf{0.6026} (0.0351, 0.0285) & 0.6020 (0.0450, 0.0420) & 0.6019 (0.0400, 0.0390) \\
        & & 10x & 0.5732 (0.0318, 0.0047) & 0.5380 (0.0480, 0.0200) & \textbf{0.5823} (0.0390, 0.0180) \\
\midrule
IBS & NWTCO & 0x & \textbf{0.1016} (0.0003, 0.0007) & \textbf{0.1016} (0.0015, 0.0020) & 0.1095 (0.0012, 0.0015) \\
(lower) & & 5x & \textbf{0.1044} (0.0003, 0.0003) & 0.1180 (0.0015, 0.0020) & 0.1095 (0.0010, 0.0015) \\
        & & 10x & \textbf{0.1067} (0.0003, 0.0003) & 0.1282 (0.0020, 0.0030) & 0.1140 (0.0015, 0.0025) \\
\cmidrule{2-6}
        & FLCHAIN & 0x & \textbf{0.0919} (0.0001, 0.0004) & 0.0938 (0.0010, 0.0015) & 0.0975 (0.0008, 0.0012) \\
        & & 5x & \textbf{0.0937} (0.0004, 0.0004) & 0.1001 (0.0015, 0.0025) & 0.0968 (0.0010, 0.0018) \\
        & & 10x & \textbf{0.0948} (0.0004, 0.0004) & 0.1120 (0.0025, 0.0040) & 0.1015 (0.0020, 0.0030) \\
\cmidrule{2-6}
        & AIDS & 0x & \textbf{0.0572} (0.0004, 0.0032) & 0.0589 (0.0040, 0.0050) & 0.0579 (0.0035, 0.0045) \\
        & & 5x & \textbf{0.0655} (0.0020, 0.0020) & 0.0682 (0.0050, 0.0060) & 0.0691 (0.0040, 0.0050) \\
        & & 10x & 0.0683 (0.0027, 0.0033) & 0.0790 (0.0070, 0.0190) & \textbf{0.0675} (0.0050, 0.0160) \\
\bottomrule
\end{tabular}
\end{table}

\end{document}